\documentclass[10pt,twocolumn,letterpaper]{article}

\usepackage{cvpr}
\usepackage{times}
\usepackage{epsfig}
\usepackage{graphicx}
\usepackage{amsmath}
\usepackage{amssymb}
\usepackage{multirow}

\usepackage[pagebackref=true,breaklinks=true,letterpaper=true,colorlinks,bookmarks=false]{hyperref}
\cvprfinalcopy % *** Uncomment this line for the final submission

\def\cvprPaperID{121} % *** Enter the CVPR Paper ID here

\ifcvprfinal\fi
\begin{document}
	
	%%%%%%%%% TITLE
	\title{Instance Embedding Transfer to Unsupervised Video Object Segmentation}
	\author{Siyang Li$^{1}$, Bryan Seybold$^{2}$, Alexey Vorobyov$^{2}$, Alireza Fathi$^{2}$, Qin Huang$^{1}$, and C.-C. Jay Kuo$^{1}$\\
		$^{1}$University of Southern California, $^{2}$Google Inc.\\
		{\tt\small siyangl@usc.edu, \{seybold,vorobya,alirezafathi\}@google.com,}\\ {\tt\small qinhuang@usc.edu, cckuo@sipi.usc.edu}
	}
	% \author{Siyang Li\\
	% University of Southern California\\
	% %Institution1 address\\
	% {\tt\small siyangl@usc.edu}
	% \and
	% Bryan Seybold\\
	% Google Inc.\\
	% %First line of institution2 address\\
	% {\tt\small seybold@google.com}
	% \and
	% Alexey Vorobyov\\
	% Google Inc.\\
	% %First line of institution2 address\\
	% {\tt\small vorobya@google.com }
	% \and
	% Alireza Fathi\\
	% Google Inc.\\
	% {\tt\small alirezafathi@google.com}
	% %First line of institution2 address\\
	% \and
	% Qin Huang\\
	% University of Southern California\\
	% %Institution1 address\\
	% {\tt\small qinhuang@usc.edu}
	% \and
	% C.-C. Jay Kuo\\
	% University of Southern California\\
	% %Institution1 address\\
	% {\tt\small cckuo@sipi.usc.edu}
	% }
	
	\maketitle
	%\thispagestyle{empty}
	
	%%%%%%%%% ABSTRACT
	\begin{abstract}
		We propose a method for unsupervised video object segmentation by transferring the knowledge encapsulated in image-based instance embedding networks. The instance embedding network produces an embedding vector for each pixel that enables identifying all pixels belonging to the same object. Though trained on static images, the instance embeddings are stable over consecutive video frames, which allows us to link objects together over time. Thus, we adapt the instance networks trained on static images to video object segmentation and incorporate the embeddings with objectness and optical flow features, without model retraining or online fine-tuning. The proposed method outperforms state-of-the-art unsupervised segmentation methods in the DAVIS dataset and the FBMS dataset.
	\end{abstract}
	
	%%%%%%%%% BODY TEXT
	\section{Introduction}\label{sec:intro}
	One important task in video understanding is object localization in time and space. Ideally, it should be able to localize familiar or novel objects consistently over time with a sharp object mask, which is known as video object segmentation (VOS). If no indication of which object to segment is given, the task is known as unsupervised video object segmentation or primary object segmentation. Once an object is segmented, visual effects and video understanding tools can leverage that information~\cite{ARsurvey, videocomposition}. 
	
	Related object segmentation tasks in static images are currently dominated by methods based on the fully convolutional neural network (FCN) ~\cite{DeepLab, FCN}. These neural networks require large datasets of segmented object images such as PASCAL~\cite{Pascal2010} and COCO~\cite{COCO2014}. Video segmentation datasets are smaller because they are more expensive to annotate~\cite{segtrack, fbms59, Perazzi2016}. As a result, it is more difficult to train a neural network explicitly for video segmentation. Classic work in video segmentation produced results using optical flow and shallow appearance models~\cite{kohprimary, msg, key, FST, GrundmannKwatra2010, wang2015saliency} while more recent methods typically pretrain the network on image segmentation datasets and later adapt the network to the video domain, sometimes combined with optical flow~\cite{OSVOS, LMP, segflow, OnAVOS, tokmakov2017learning, FSEG}. 
	
	\begin{figure}
		\centering
		\includegraphics[width=0.32\columnwidth]{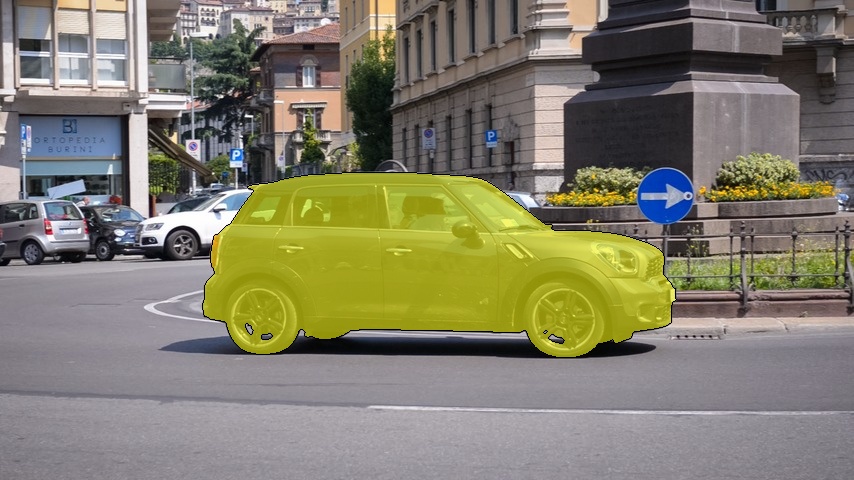}
		\includegraphics[width=0.32\columnwidth]{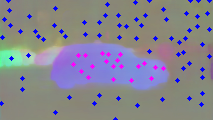}
		\includegraphics[width=0.32\columnwidth]{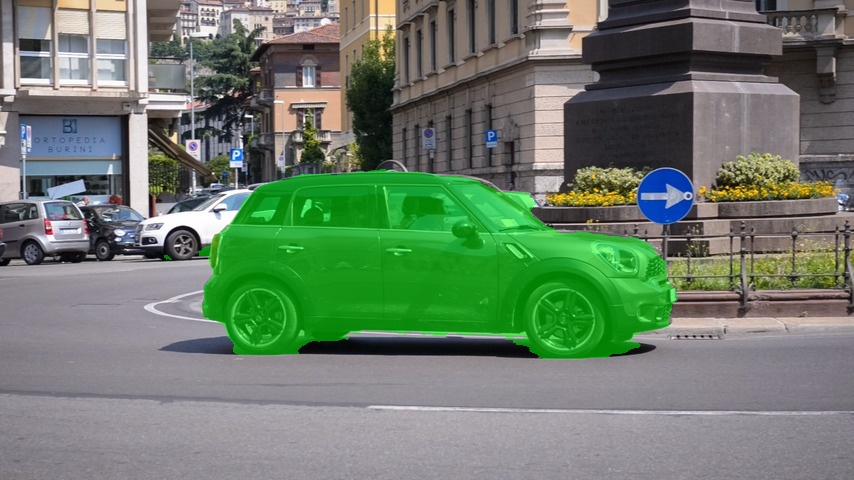}\\
		\includegraphics[width=0.32\columnwidth]{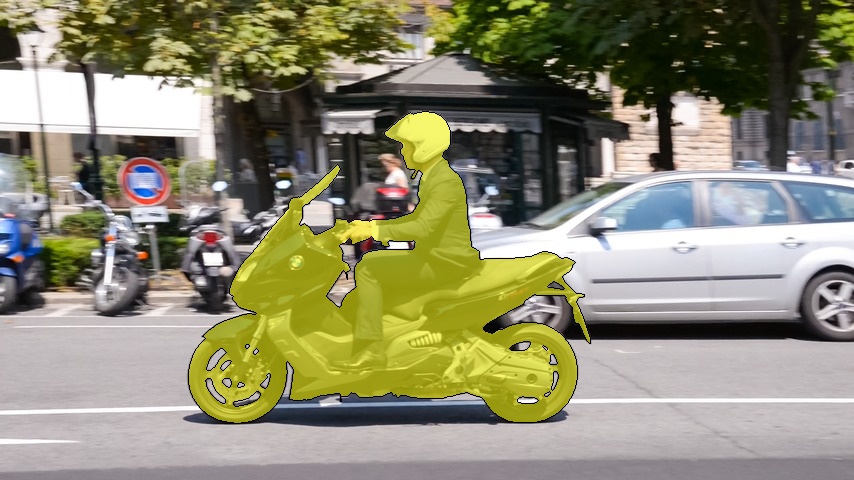}
		\includegraphics[width=0.32\columnwidth]{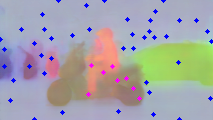}
		\includegraphics[width=0.32\columnwidth]{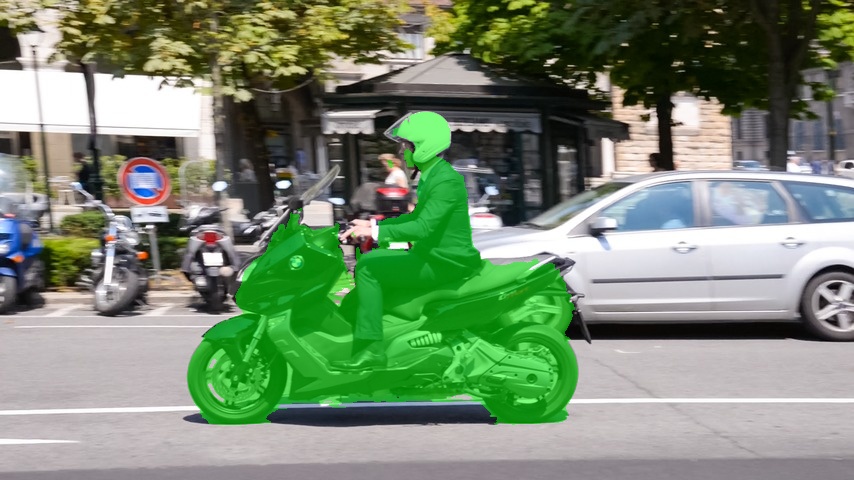}
		\caption{An example of the changing segmentation target (foreground) in videos depending on motion. A car is the foreground in the top video while a car is the background in the bottom video. To address this issue, our method obtains embeddings for object instances and identifies representative embeddings for foreground/background and then segments the frame based on the representative embeddings. 
			\textbf{Left}: the ground truth. 
			\textbf{Middle}: A visualization of the embeddings projected into RGB space via PCA, along with representative points for the foreground (magenta) and background (blue).
			\textbf{Right}: the segmentation masks produced by the proposed method. Best viewed in color.}
		\label{fig:example_intro}
		\vspace{-0.5cm}
	\end{figure}
	
	In this paper, we propose a method to transfer the knowledge encapsulated in instance segmentation embeddings learned from static images and integrate it with objectness and optical flow to segment a moving object in video. Instead of training an FCN that directly classifies each pixel as foreground/background as in~\cite{tokmakov2017learning, FSEG, segflow, LMP}, we train an FCN that jointly learns object instance embeddings and semantic categories from images~\cite{fathi2017semantic}. The distance between the learned embeddings encodes the similarity between pixels. We argue that the instance embedding is a more useful feature to transfer from images to videos than a foreground/background prediction. As shown in Fig.~\ref{fig:example_intro}, cars appear in both videos but belong to different categories (foreground in the first video and background in the second video). If the network is trained to directly classify cars as foreground on the first video, it tends to classify the cars as foreground in the second video as well. As a result, the network needs to be fine-tuned for each sequence~\cite{OSVOS}. In contrast, the instance embedding network can produce unique embeddings for the car in both sequences without interfering with other predictions or requiring fine-tuning. The task then becomes selecting the correct embeddings to use as an appearance model.
	% Previous methods address this problem by leveraging the motion information~\cite{LMP, tokmakov2017learning, segtrack} for training, but most of those methods still require expensive video segmentation annotations. In our method, the instance embedding network is decoupled from motion, allowing training on images alone.
	Relying on the embeddings to encode object instance information, we propose a method to identify the representative embeddings for the foreground (target object) and the background based on objectness scores and optical flow. Visual examples of the representative embeddings are displayed in the middle column of Fig.~\ref{fig:example_intro}. Finally, all pixels are classified by finding the nearest neighbor in a set of representative foreground or background embeddings. This is a non-parametric process requiring no video specific supervision for training or testing. 
	
	We evaluate the proposed method on the DAVIS dataset~\cite{Perazzi2016} and the FBMS dataset~\cite{fbms59}. Without fine-tuning the embedding network on the target datasets, we obtain better performance than previous state-of-the-art methods. More specifically, we achieve a mean intersection-over-union (IoU) of 78.5\% and 71.9\% on the DAVIS dataset~\cite{Perazzi2016} and the FBMS dataset~\cite{fbms59}, respectively. 
	
	To summarize, our main contributions include
	\vspace{-0.2cm}
	\begin{itemize}
		\setlength{\itemsep}{0pt}
		\setlength{\parskip}{0pt}
		\setlength{\parsep}{0pt}
		\item A new strategy for adapting instance segmentation models trained on static images to videos. Notably, this strategy performs well on video datasets without requiring any video object segmentation annotations. 
		
		\item This strategy outperforms previously published unsupervised methods on both DAVIS benchmark and FBMS benchmark and approaches the performance of semi-supervised CNNs without requiring retraining any networks at test time. 
		
		\item Proposal of novel criteria for selecting a foreground object without supervision, based on semantic score and motion features over a track. 
		
		\item Insights into the stability of instance segmentation embeddings over time. 
	\end{itemize}
	
	\section{Related Work}
	\noindent\textbf{Unsupervised video object segmentation.}
	Unsupervised video object segmentation discovers the most salient, or primary, objects that move against a video's background or display different color statistics. One set of methods to solve this task builds hierarchical clusters of pixels that may delineate objects~\cite{GrundmannKwatra2010}. Another set of methods performs binary segmentation of foreground and background. Early foreground segmentation methods often used Gaussian Mixture Models and Graph Cut~\cite{BVS, ObjectFlow}, but more recent work uses convolutional neural networks (CNN) to identify foreground pixels based on saliency, edges, and/or motion ~\cite{LMP, tokmakov2017learning, FSEG}. For example, LMP~\cite{LMP} trains a network which takes optical flow as an input to separate moving and non-moving regions and then combines the results with objectness cues from SharpMask~\cite{sharpmask} to generate the moving object segmentation. LVO~\cite{tokmakov2017learning} trains a two-stream network, using RGB appearance features and optical flow motion features that feed into a ConvGRU~\cite{convgru} layer to generate the final prediction. FSEG~\cite{FSEG} also proposes a two-stream network trained with mined supplemental data. SfMNet~\cite{sfmnet} uses differentiable rendering to learn object masks and motion models without mask annotations. Despite the risk of focusing on the wrong object, unsupervised methods can be deployed in more places because they do not require user interaction to specify an object to segment. Since we are interested in methods requiring no user interaction, we choose to focus on unsupervised segmentation.
	
	\noindent\textbf{Semi-supervised video object segmentation.} 
	Semi-supervised video object segmentation utilizes human annotations on the first frame of a video (or more) indicating which object the system should track. Importantly, the annotation provides a very good appearance model initialization that unsupervised methods lack. The problem can be formulated as either a binary segmentation task conditioned on the annotated frame or a mask propagation task between frames. Non-CNN methods typically rely on Graph Cut~\cite{BVS, ObjectFlow}, but CNN based methods offer better accuracy ~\cite{MSK, OSVOS, OnAVOS, segflow, CTN, PLM}. Mask propagation CNNs take in the previous mask prediction and a new frame to propose a segmentation in the new frame. VPN~\cite{VPN} trains a bilateral network to propagate to new frames. MSK~\cite{MSK} trains a propagation network with synthetic transformations of still images and applies the same technique for online fine-tuning. SegFlow~\cite{segflow} finds that jointly learning moving object masks and optical flow helps to boost the segmentation performance. Binary segmentation CNNs typically utilize the first frame for fine-tuning the network to a specific sequence. The exact method for fine-tuning varies: OSVOS~\cite{OSVOS} simply fine-tunes on the first frame. OnAVOS~\cite{OnAVOS} fine-tunes on the first frame and a subset of predictions from future frames. Fine-tuning can take seconds to minutes, and longer fine-tuning typically results in better segmentation. Avoiding the time cost of fine-tuning is a further inducement to focus on unsupervised methods. 
	
	\noindent\textbf{Image segmentation.}
	Many video object segmentation methods~\cite{OnAVOS, OSVOS, MSK} are built upon image semantic segmentation neural networks~\cite{FCN, DeepLab, RAN}, which predict a category label for each pixel. These fully convolutional networks allow end-to-end training on images of arbitrary sizes. Semantic segmentation networks do not distinguish different instances from the same object category, which limits their suitability to video object segmentation. Instance segmentation networks~\cite{fathi2017semantic, DengInstanceEmbedding, MaskRCNN} can label each instance uniquely. Instance embedding methods~\cite{fathi2017semantic, DengInstanceEmbedding, de2017semantic} provide an embedding space where pixels belonging to the same instance have similar embeddings. Spatial variations in the embeddings indicate the edges of object masks. Relevant details are given in Sec.~\ref{sec:features}. It was unknown if instance embeddings are stable over time in videos, but we hypothesized that these embeddings might be useful for video object segmentation. 
	
	\begin{figure*}
		\includegraphics[width=\textwidth]{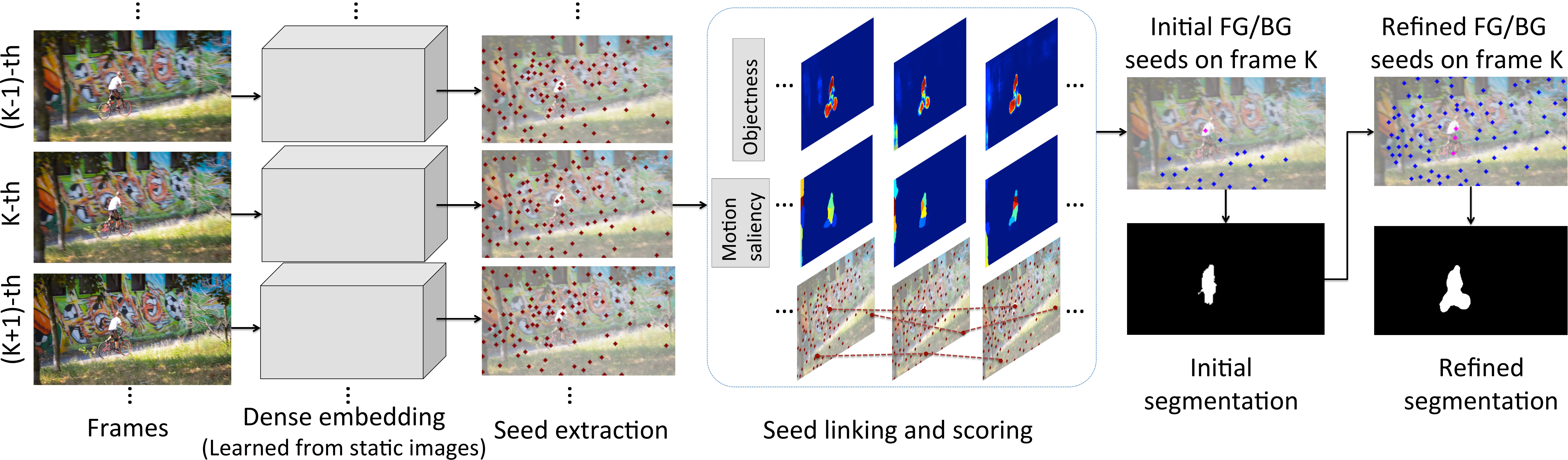}
		\caption{An overview of the proposed method. Given the video sequences, the dense embeddings are obtained by applying an instance segmentation network trained on static images. Then representative embeddings, called seeds, are obtained. Seeds are linked across the whole sequence (we show 3 consecutive frames as an illustration here). The seed with the highest score based on objectness and motion saliency is selected as the initial seed (in magenta) to grow the initial segmentation. Finally, more foreground seeds as well as background seeds are identified to refine the segmentation.}\label{fig:overview}
		\vspace{-0.5cm}
	\end{figure*}
	
	\section{Proposed Method}
	%\subsection{Overview}
	An overview of the proposed method is depicted in Fig.~\ref{fig:overview}. We first obtain instance embeddings, objectness scores, and optical flow that we will use as inputs (Sec.~\ref{sec:features}). Based on the instance embeddings, we identify ``seed'' points that mark segmentation proposals (Sec.~\ref{sec:seed_proposals}). Consistent proposal seeds are linked to build seed tracks, and we rank the seed tracks by objectness scores and motion saliency to select a foreground proposal seed on every frame (Sec.~\ref{sec:seed_ranking}). We further build a set of foreground/background proposal seeds to produce the final segmentation mask in each frame (Sec.~\ref{sec:seed_refinement}).

	\subsection{Extracting features}\label{sec:features}
	Our method utilizes three inputs: instance embeddings, objectness scores, and optical flow. None of these features are fine-tuned on video object segmentation datasets or fine-tuned online to specific sequences. The features are extracted for each frame independently as follows.
	
	\noindent\textbf{Instance embedding and objectness.} We train a network to output instance embeddings and semantic categories on the image instance segmentation task as in \cite{fathi2017semantic}. Briefly, the instance embedding network is a dense-output convolutional neural network with two output heads trained on static images from an instance segmentation dataset.
	
	The first head outputs an embedding for each pixel, where pixels from same object instance have smaller Euclidean distances between them than pixels belonging to separate objects. Similarity $R$ between two pixels $i$ and $j$ is measured as a function of the Euclidean distance in the $E$-dimensional embedding space, $\mathbf{f}$,
	\begin{align}\label{eq:sim}
	R(i, j) = \frac{2}{1 + \exp(||\mathbf{f}(i) - \mathbf{f}(j)||_{2}^2)}.
	\end{align}
	
	This head is trained by minimizing the cross entropy between the similarity and the ground truth matching indicator $g(i, j)$. For locations $i$ and $j$, the ground truth matching indicator $g(i, j) = 1$ if pixels belong to the same instance and $g(i, j) = 0$ otherwise. The loss is given by
	\begin{equation}\label{eq:sim_loss}
	\begin{split}
	L_s = -\frac{1}{|A|}\sum_{i, j\in A}&w_{ij}[g(i, j)\log(R(i, j))\\
	&+ (1-g(i, j))\log(1-R(i, j))],
	\end{split}
	\end{equation}
	where $A$ is a set of pixel pairs, $R(i,j)$ is the similarity between pixels $i$ and $j$ in the embedding space and $w_{ij}$ is inversely proportional to instance size to balance training.
	
	The second head outputs an objectness score from semantic segmentation. We minimize a semantic segmentation log loss to train the second head to output a semantic category probability for each pixel. The objectness map is derived from the semantic prediction as
	\begin{align}\label{eq:objectness}
	O(i) = 1 - P_{BG}(i),
	\end{align}
	where $P_{BG}(i)$ is the probability that pixel $i$ belongs to the semantic category ``background''\footnote{Here in semantic segmentation, ``background'' refers to the region that does not belong to any category of interest, as opposed to video object segmentation where the ``background'' is the region other than the target object. We use ``background'' as in video object segmentation for the rest of the paper.}. We do not use the scores for any class other than the background in our work.
	
	\noindent\textbf{Embedding graph.} We build a 4-neighbor graph from the dense embedding map, where each embedding vector becomes a vertex and edges exist between spatially neighboring embeddings with weights equal to the Euclidean distance between embedding vectors. This embedding graph will be used to generate image regions later. A visualized embedding graph is shown in Fig.~\ref{fig:instance_edge}.
	
	\noindent\textbf{Optical flow.} The motion saliency cues are built upon optical flow. For fast optical flow estimation at good precision, we utilize a reimplementation of FlowNet 2.0~\cite{flownet}, an iterative neural network.
	
	\subsection{Generating proposal seeds}\label{sec:seed_proposals}
	We propose a small number of representative seed points $S^k$ in frame $k$ for some subset of frames $K$ (typically all) in the video. Most computations only compare against seeds within the current frame, so the superscript $^k$ is omitted for clarity unless the computation is across multiple frames. The seeds we consider as FG or BG should be diverse in embedding space because the segmentation target can be a moving object from an arbitrary category. In a set of diverse seeds, at least one seed should belong to the FG region. We also need at least one BG seed because the distances in the embedding space are relative. The relative distances in embedding space, or similarity from Eq.~\ref{eq:sim}, from each point to the FG and BG seed(s) can be used to assign a labels to all pixels.

	\noindent\textbf{Candidate points.}
	In addition to being diverse, the seeds should be representative of objects. The embeddings on the boundary of two objects are usually not close to the embedding of either object. Because we want embeddings representative of objects, we exclude seeds from object boundaries. To avoid object boundaries, we only select seeds from candidate points where the instance embeddings are locally consistent. (An alternative method to identify the boundaries to avoid would be to use an edge detector such as \cite{SED, HED}.) We construct a map of embedding edges by mapping discontinuities in the embedding space. The embedding edge map is defined as the ``inverse'' similarity in the embedding space within the neighbors around each pixel,
	\begin{align}
	c(p) = \max_{q\in N(p)} 1 - R(p, q),
	\end{align}
	where $p$ and $q$ are pixel locations, $N(p)$ contains the four neighbors of $p$, and $R(r,q)$ is the similarity measure given in Eq.~\ref{eq:sim}. Then in the edge map we identify the pixels which are the minimum within a window of $n\times n$ centered at itself. These pixels from the candidate set $C$. Mathematically,
	\begin{align}
	C = \{p|c(p) = \min_{q\in W(p)} c(q)\},
	\end{align}
	where $W(p)$ denotes the local window.
	% $W(p)$ are pixels within an $n\times n$ window of $p$.
	% SIYANG: Sorry I didn't make it clear.
	% The edge map c(p) is obtained by only looking at its 4 neighbors. Then I apply the minimum filter with window size parameter=n to the edge map and compare the filtered map with the original edge map. The pixels have equal value in the maps will be identified as local minima.
	% Thanks for catching this change. That is different.
	
\begin{figure}
		\centering
		\begin{minipage}[b]{0.32\columnwidth}
			\centering
			\includegraphics[width=\columnwidth]{figures/motion_saliency/00034.jpg}\\
			\footnotesize{Original image}
		\end{minipage}
		\begin{minipage}[b]{0.64\columnwidth}
			\centering
			\includegraphics[width=\columnwidth]{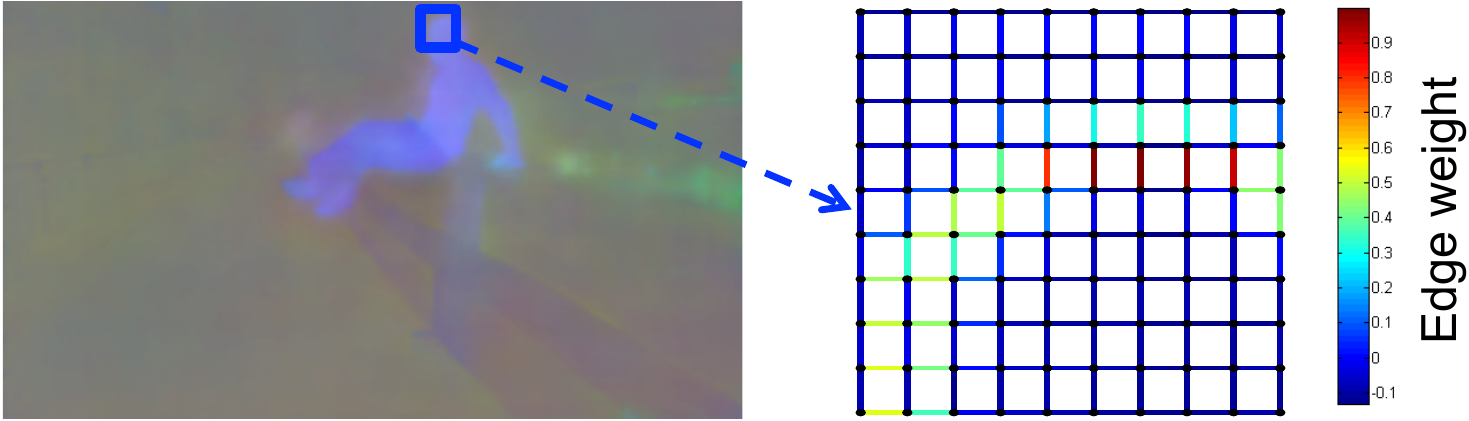}\\
			\footnotesize{Embedding graph}
		\end{minipage}\\
		\vspace{0.1cm}
		\begin{minipage}[t]{0.32\columnwidth}
			\centering
			\includegraphics[width=\columnwidth]{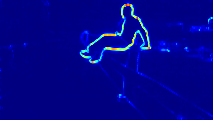}\\
			\footnotesize{Embedding edge map}
		\end{minipage}
		\begin{minipage}[t]{0.32\columnwidth}
			\centering
			\includegraphics[width=\columnwidth]{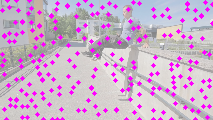}\\
			\footnotesize{Candidate set $C$}
		\end{minipage}
		\begin{minipage}[t]{0.32\columnwidth}
			\centering
			\includegraphics[width=\columnwidth]{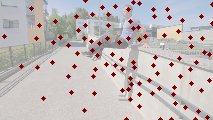}\\
			\footnotesize{Seed set $S$}
		\end{minipage}
		\vspace{0.1cm}
		\caption{\textbf{Top:} An image (\textit{left}) and the visualization of its embedding graph in the $10\times 10$ box in blue. The edge colors on the \textit{right} reflect distances between the embeddings at each pixel (the \textit{center} subfigure visualizes the embeddings via PCA). High costs appear along object boundaries.
			\textbf{Bottom:} The embedding edge map (\textit{left}), the candidate set $C$ (\textit{center}) and the seed set $S$ (\textit{right}). Best viewed in color.}
		\label{fig:instance_edge}
		\vspace{-0.5cm}
	\end{figure}

	\noindent\textbf{Diverse seeds.} These candidate points, $C$, are diverse, but still redundant with one another. We take a diverse subset of these candidates as seeds by adopting the sampling procedure from KMeans++ initialization \cite{arthur2007kmeanspp}. We only need diverse sampling rather than cluster assignments, so we do not perform the time-consuming KMeans step afterwards. The sampling procedure begins by adding the candidate point with the largest objectness score, $O(i)$, to the seed set, $S$. Sampling continues by iteratively adding the candidate, $s_{n+1}$, with smallest maximum similarity to all previously selected seeds and stops when we reach $N_{S}$ seeds,
	\begin{align}
	s_{n+1} = \arg\min_{i \in C }\max_{j\in S} R(i, j).
	\end{align}
	We repeat this procedure to produce the seeds for each frame independently, forming a seed set $S$. Note that the sampling strategy differs from \cite{fathi2017semantic}, where they consider a weighted sum of the embedding distance and semantic scores. We do not consider the semantic scores because we want to have representative embeddings for all regions of the current frame, including the background, while in \cite{fathi2017semantic}, the background is disregarded. Fig.~\ref{fig:instance_edge} shows one example of the visualized embedding edge map, the corresponding candidate set and the selected seeds.
	
	% BS QUESTION: To confirm, do you use different distances for making motion regions and for spreading out the foreground and background regions? When constructing the motion regions, I think you use the geodesic path cost, but when spreading the foreground region, your distance is the maximum link the the geodesic path? Have you tried using the same cost in both places? I would be easier to explain.
	% SIYANG: Motion saliency is based on Euclidean distance between flow vector. I agree that using geodesic distance is easier. I'll try it. % BS: please change my text if it's Euclidean distance of the flow vector. I'm not entirely sure what you mean by that.
	
	\subsection{Ranking proposed seeds}\label{sec:seed_ranking}
	In the unsupervised video object segmentation problem, we do not have an explicitly specified target object. Therefore, we need to identify a moving object as the segmentation target (i.e., FG). We first score the seeds based on objectness and motion saliency. To find the most promising seed for FG, we then build seed tracks by group embedding-consistent seeds across frames into seed tracks and aggregate scores along tracks. The objectness score is exactly $O(s)$ in Eq.~\ref{eq:objectness} for each seed. The motion saliency as well as seed track construction and ranking are explained below.
	
\begin{figure}
		\centering
		\begin{minipage}{0.32\columnwidth}
			\centering
			\includegraphics[width=\columnwidth]{figures/motion_saliency/00034.jpg}
			\footnotesize{Original image}
		\end{minipage}
		\begin{minipage}{0.32\columnwidth}
			\centering
			\includegraphics[width=\columnwidth]{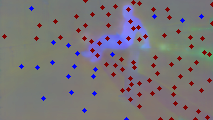}
			\footnotesize{Embedding map}
		\end{minipage}
		\begin{minipage}{0.32\columnwidth}
			\centering
			\includegraphics[width=\columnwidth]{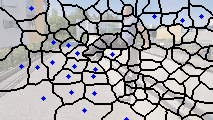}
			\footnotesize{Image regions}
		\end{minipage}\\
		\vspace{0.1cm}
		\begin{minipage}{0.32\columnwidth}
			\centering
			\includegraphics[width=\columnwidth]{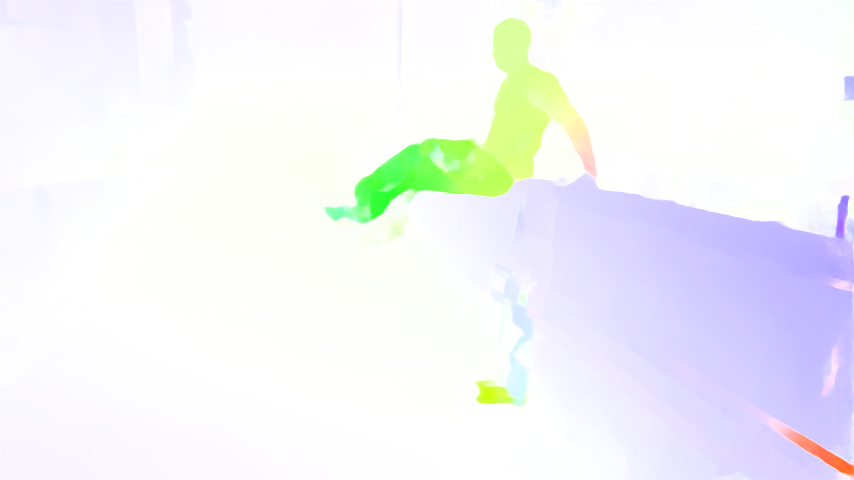}
			\footnotesize{Pixel optical flow}
		\end{minipage}
		\begin{minipage}{0.32\columnwidth}
			\centering
			\includegraphics[width=\columnwidth]{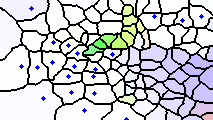}
			\footnotesize{Region optical flow}
		\end{minipage}
		\begin{minipage}{0.32\columnwidth}
			\centering
			\includegraphics[width=\columnwidth]{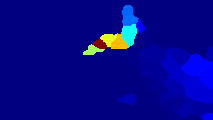}
			\footnotesize{Motion saliency}
		\end{minipage}
		\vspace{0.1cm}
		\caption{\textbf{Top:} \textit{left:} An image. \textit{center:} A projection of the embedding map into RBG space (via PCA) with the initial background seeds $S_{BG}$ marked in blue and other seeds in red. \textit{right:} The regions near each seed in the embedding graph. 
			\textbf{Bottom:} \textit{left:} The optical flow. \textit{center:} Average flow within each region. \textit{right:} A map of motion saliency scores. Best viewed in color.}
		\label{fig:motion_saliency}
	\end{figure}

	\noindent\textbf{Motion saliency.} Differences in optical flow can separate objects moving against a background~\cite{broxmalik}. Because optical flow estimation is still imperfect~\cite{flownet}, we average flow within the image regions rather than using the flow from a single pixel. The region corresponding to each seed consists of the pixels in the embedding graph from Sec.~\ref{sec:features} with the shortest geodesic distance to that seed. For each seed $s$, we use the average optical flow in the corresponding region as $\mathbf{v}_s$. An example of image regions is shown in Fig.~\ref{fig:motion_saliency}.
	
	% AV QUESTION: It seems to me that definition of the region is similar to the definition of Seed similarity graph in corresponding section. % BS: they are similar. I'd love a better way to talk about these, but the distance measure in the graph changes at different points and I'm not sure how to talk about that. Siyang, do you have any ideas for how to simplify this text? (Also, please make sure that it's still accurate after I modified it.)
	
	Then we construct a model of the background. First, $N_{BG}$ seeds with the lowest objectness score, $O(s)$, are selected as the initial background seeds, denoted by $S_{BG}$. The set of motion vectors associated with these seeds forms our background motion model $V_{BG}$. The motion saliency for each seed, $s$, is the normalized distance to the nearest background motion vector,
	\begin{align}
	M(s) = \frac{1}{Z}\min_{\mathbf{v}_b \in V_{BG}}||\mathbf{v}_s - \mathbf{v}_b||_{2}^{2},
	\end{align}
	where the normalization factor $Z$ is given by
	\begin{align}
	Z=\max_{s \in S}(\min_{\mathbf{v}_b\in V_{BG}}||\mathbf{v}_s - \mathbf{v}_b||_{2}^{2}).
	\end{align}
	There are other approaches to derive motion saliency from optical flow. For example, in \cite{kohprimary}, motion edges are obtained by applying some edge detector to optical flow and then motion saliency of some region is computed as a function of motion edge intensity. The motion saliency proposed in this work is more efficient and works well in terms of the final segmentation performance.
	
	\noindent\textbf{Seed tracks.}
	Another property of the foreground object is that it should be a salient object in multiple frames. We score this by linking similar seeds together across frames into a seed track and taking the average product of objectness and motion saliency scores over each track. The $j$-th seed on frame 0, $s^{0}_{j}$, initiates a seed track, $T_{j}$. $T_{j}$ is extended frame by frame by adding the seed with the highest similarity to $T_{j}$. Specifically, supposing that we have a track $T_{j}^{m}$ across frames 0-$m$, it is extended to frame $m+1$ by adding the most similar seed on frame $m+1$ to $T_{j}^{m}$, forming $T_{j}^{m+1}$:
	\begin{align}
	r = \arg\max_{s \in S^{m+1}}\sum_{t \in T_{j}^{m}} R(s, t),
	\end{align}
	where $R(s, t)$ is the similarity measure given by Eq.~\ref{eq:sim}, and $r$ is the seed in frame $m+1$ with the highest similarity to $T_{j}^{m}$. Then we have
	$T_{j}^{m+1} = T_{j}^{m}\bigcup \{r\}$.
	Eventually, we have $T_{j}$ starting from $s_{j}^{0}$ and ending at some seed on the last frame.
	The foreground score for $T_{j}$ is
	\begin{align}\label{eq:fg_ranking}
	F(T_{j}) = \frac{1}{|T_{j}|}\sum_{s\in T_{j}}O(s)M(s),
	\end{align}
	where $|T_{j}|$ is the size of the seed track, equal to the sequence length. 
	
	\subsection{Segmenting a foreground proposal}\label{sec:seed_refinement}
	
	% BS: How much latency does this save? It seems like focusing on a negative. I think the only added latency is the cost of optical flow, so I'm not sure why this produces long latency.
	% SIYANG: In order to determine the init seeds on the first frame, the current algorithm needs to wait until embeddings of all frames are computed. I'm not sure about the optical flow. Does the stablized flow computation need the unstablized OF for the whole sequence? If that's true, then what I dicussed here doesn't change much. I'll delete that part.
	% BS: I'm going to recommend removing this discussion to save space, but we could keep it in if you feel strongly about it.
	
	% The straightforward foreground seed selection scheme is conducted when the embeddings and optical flow are available for all frames, resulting in long latency. We discuss some varients to reduce the latency in Sec.~\ref{sec:ablation}.

	\noindent\textbf{Initial foreground segmentation.}
	The seed track with the highest foreground score is selected on each frame to provide an initial foreground seed, denoted by $s_{FG}^{k}$. We obtain an initial foreground segmentation by identifying pixels close to the foreground seed $s_{FG}^{k}$ in the embedding graph explained in Sec.~\ref{sec:features}. Here the distance, denoted by $d(p,s)$, between any two nodes, $p$ and $s$, is defined as the maximum edge weight along the shortest geodesic path connecting them. We again take the $N_{BG}$ seeds with the lowest objectness scores as the initial background seed set, $S_{BG}^k$. Then the initial foreground region $I_{FG}$ is composed of the pixels closer to the foreground seed $s_{FG}^{k}$ than any background seeds,
	%As shown in Fig.~\ref{fig:expand}, we recompute geodesic distances in the embedding graph as the maximum edge weight along the shortest path between a pixel and a seed, denoted by $d(p, s)$. 
	\begin{align}
	I_{FG} = \{p | d(p, s_{FG}^k) < \min_{b\in S_{BG}^k} d(p, b)\}.
	\end{align}
	% BS: I'm not sure this figure was adding a lot, so perhaps we shouldn't include it? It would save us cutting more text. Thoughts? Alternatively, we might be able to push more information into this figure and reduce the amount of text.
	% \begin{figure}
	%     \centering
	%     \includegraphics[width=0.32\columnwidth]{figures/parkour/parkour_00033.jpg}
	%     \includegraphics[width=0.32\columnwidth]{figures/FG_expand.png}
	%     \includegraphics[width=0.32\columnwidth]{figures/parkour/fg_region_00033.png}
	%     \caption{The initial segmentation process. \textbf{Left}: the original frame. \textbf{Middle}: illustration of the initial foreground segmentation. For node $X$, the shortest paths to the FG embedding and BG embedding are highlighted in green and red, respectively. The largest weight on the two paths are 0.3 and 0.5 respectively. Thus node 'X' is closer to FG though the corresponding shortest path is longer than the one to background. \textbf{Right}: the initial foreground mask.}
	%     \label{fig:expand}
	% \end{figure}
	% % BS: Please change the red/green color scheme to magenta and blue to match other figures and help for colorblindness.
	
	\noindent\textbf{Adding foreground seeds.}
	Often, selecting a single foreground seed is overly conservative and the initial segmentation fails to cover an entire object. To expand the foreground segmentation, we create a set of foreground seeds, $S_{FG}^{k}$ from the combination of $s_{FG}^k$ and seeds marking image regions mostly covered by the initial foreground segmentation. These image regions are the ones described in Sec.~\ref{sec:seed_ranking} and Fig.~\ref{fig:motion_saliency}. If more than a proportion $\alpha$ of a region intersects with the initial foreground segmentation, the corresponding seed is added to the $S_{FG}^{k}$.
	
	\noindent\textbf{Adding background seeds.}
	The background contains two types of regions: non-object regions (such as sky, road, water, etc.) that have low objectness scores, and static objects, which are hard negatives because in the embedding space, they are often closer to the foreground than the non-object background. Static objects are particularly challenging when they belong to the same semantic category as the foreground object\footnote{E.g., the ``camel'' sequence in DAVIS in supplementary materials.}. We expand our representation of the BG regions by taking the union of seeds with object scores less than a threshold $O_{BG}$ and seeds with motion saliency scores less than a threshold $M_{BG}$:
	\begin{align}
	S_{BG}^{k} = \{s | O(s) \leq O_{BG}\} \bigcup \{s | M(s) \leq M_{BG}\}.
	\end{align}
	
	\noindent\textbf{Final segmentation.}
	Once the foreground $S_{FG}^{k}$ and background $S_{BG}^{k}$ sets are established, similarity to the nearest foreground and background seeds is computed for each pixel. It is possible to use the foreground and background sets from one frame to segment another frame for FG/BG similarity computation:
	\begin{align}\label{eq:distance_compute}
	R_{FG}^{l}(i^l) = \max_{s\in S^{k}_{FG}} R(i^l, s),\\
	R_{BG}^{l}(i^l) = \max_{s\in S^{k}_{BG}} R(i^l, s),
	\end{align}
	where pixels on the target frame $l$ are denoted by $i^{l}$.
	Instead of directly propagating the foreground or background label from the most similar seed to the embedding, we obtain a soft score as the confidence of the embedding $i^l$ being foreground:
	\begin{align}\label{eq:fg_prob}
	P_{FG}(i^l) = \frac{R_{FG}^{k}(i^l)}{R_{FG}^{k}(i^l) + R_{BG}^{k}(i^l)}.
	\end{align}
	
	Finally, the dense CRF~\cite{densecrf} is used to refine the segmentation mask, with the unary term set to the negative log of $P_{FG}(i^l)$, as in \cite{DeepLab}.
	
	\noindent\textbf{Online adaptation.}\label{sec:online_adaptation}
	Online adaptation of our method is straightforward: we simply generate new sets of foreground and background seeds. This is much less expensive than fine-tuning an FCN for adaptation as done in \cite{OnAVOS}. Though updating the foreground and background sets could result in segmenting different objects in different frames, it improves the results in general, as discussed in Sec.~\ref{sec:ablation}.
	
	\section{Experiments}
	
	\begin{table*}[t!h]
		\begin{center}
			\setlength\tabcolsep{5pt}
			\begin{tabular}{|c|c|c|c|c|c|c|c|c|c|}
				\hline
				&NLC~\cite{NLC} &CUT~\cite{CUT} &FST~\cite{FST} &SFL~\cite{segflow} &LVO~\cite{tokmakov2017learning} &MP~\cite{LMP} &FSEG~\cite{FSEG} & ARP\cite{kohprimary} & Ours \\
				\hline
				\small{Fine-tune on DAVIS?} & No &Yes &No &Yes  &Yes &No &No &No &No\\
				\hline
				$\mathcal{J}$ Mean &55.1 &55.2	&55.8 &67.4  &75.9 & 70.0 & 70.7 &76.2 & \textbf{78.5}\\
				\hline
				%J Recall &\\
				%\hline
				$\mathcal{F}$ Mean &52.3 &55.2 &51.1 &66.7 &72.1 &65.9 &65.3 &70.6 &\textbf{75.5}\\
				\hline
				%F recall\\
			\end{tabular}
		\end{center}
		\caption{The results on the \textit{val} set of DAVIS 2016 dataset~\cite{Perazzi2016}. Our method achieves the highest in both evaluation metrics, and outperforms the methods fine-tuned on DAVIS. Online adaptation is applied on every frame. Per-video results are listed in supplementary materials.}
		\label{table:davis}
	\end{table*}
	
	\begin{figure*}
		\centering
		\includegraphics[width=0.19\textwidth]{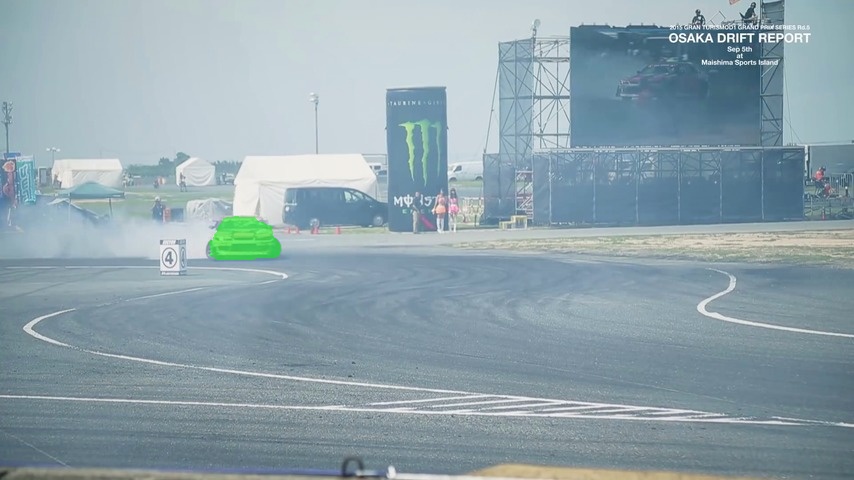}
		\includegraphics[width=0.19\textwidth]{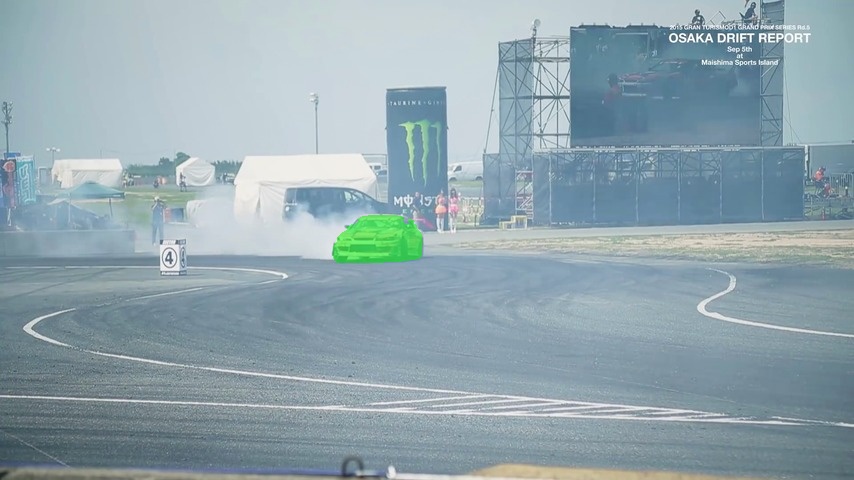}
		\includegraphics[width=0.19\textwidth]{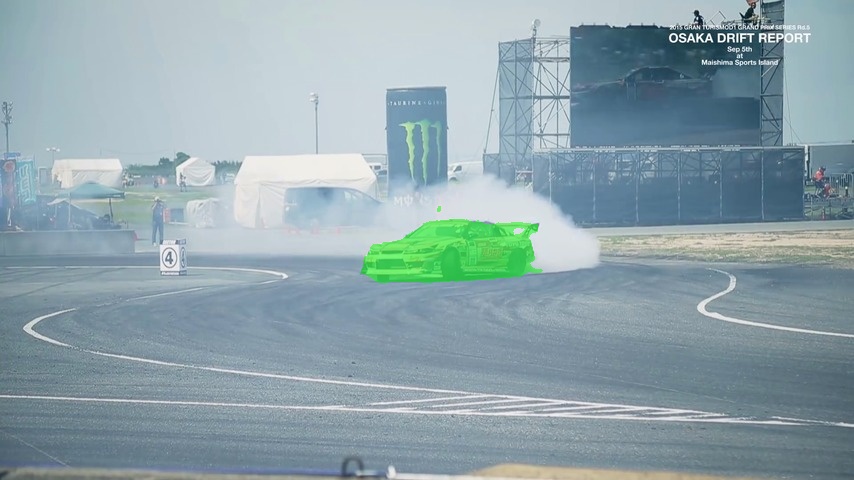}
		\includegraphics[width=0.19\textwidth]{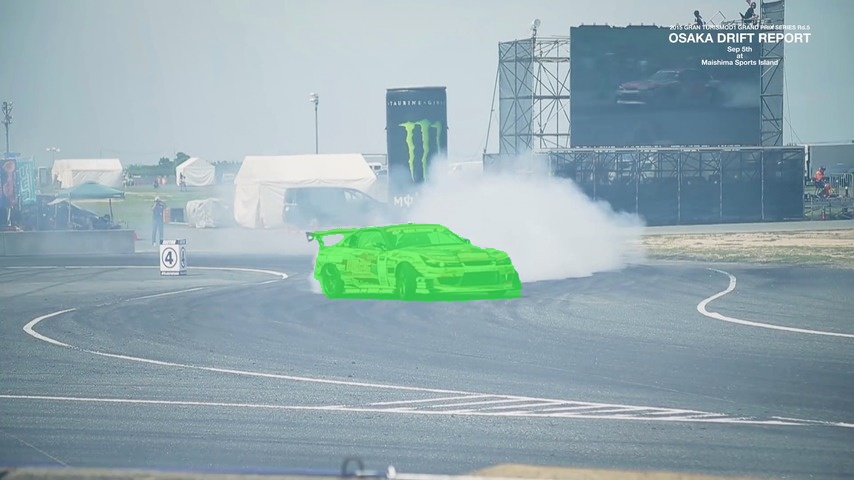}
		\includegraphics[width=0.19\textwidth]{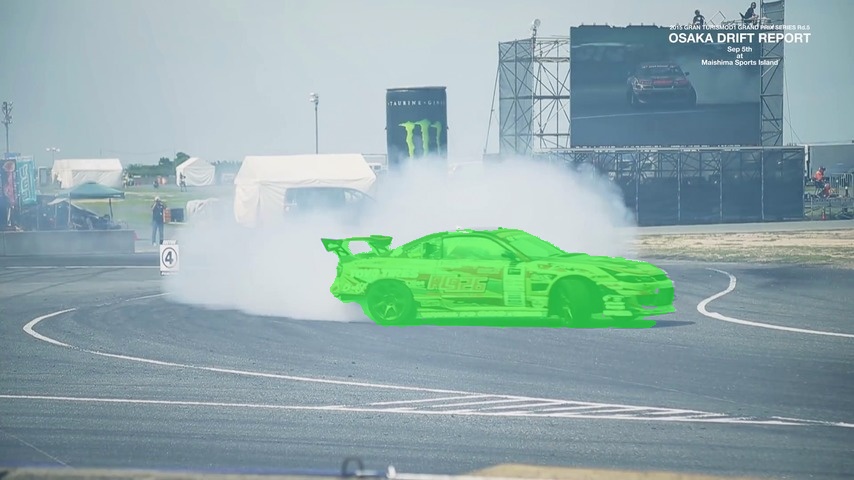}\\
		\includegraphics[width=0.19\textwidth]{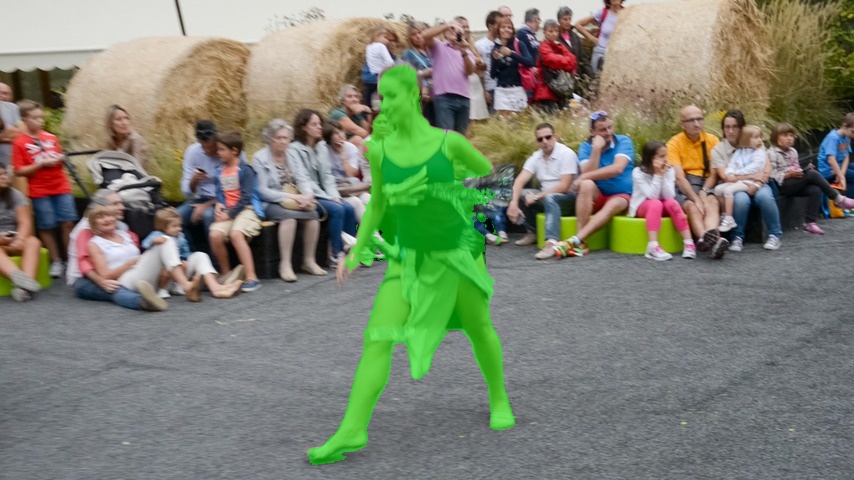}
		\includegraphics[width=0.19\textwidth]{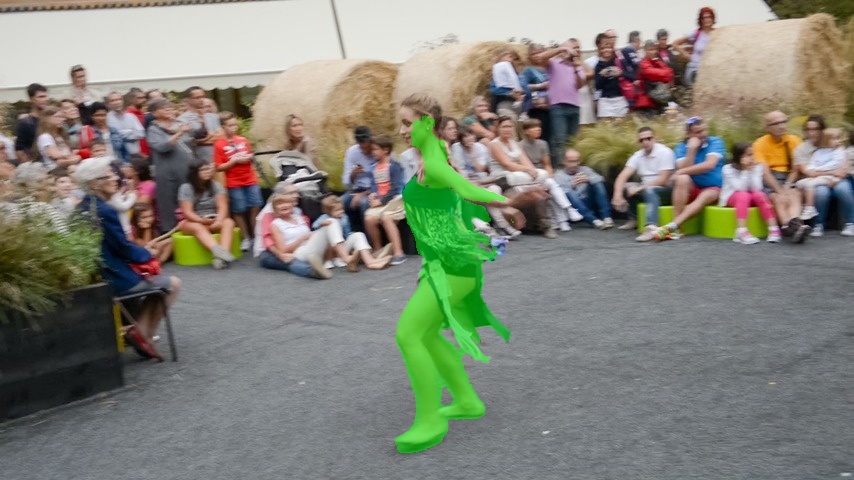}
		\includegraphics[width=0.19\textwidth]{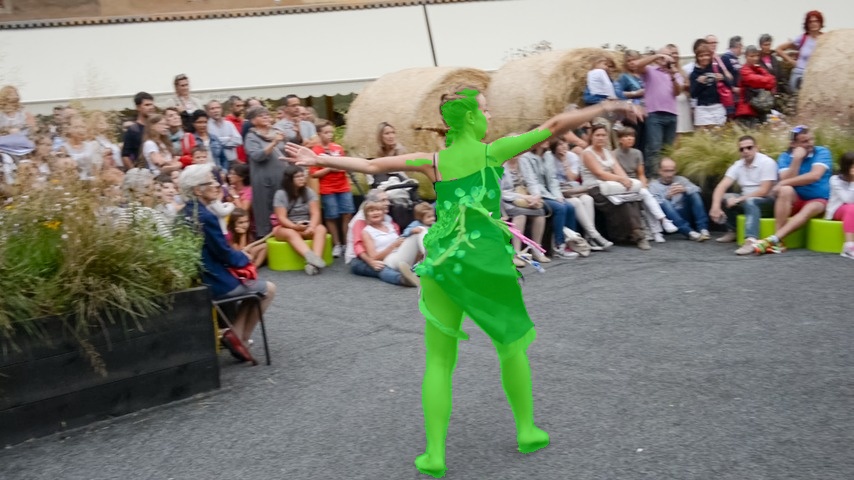}
		\includegraphics[width=0.19\textwidth]{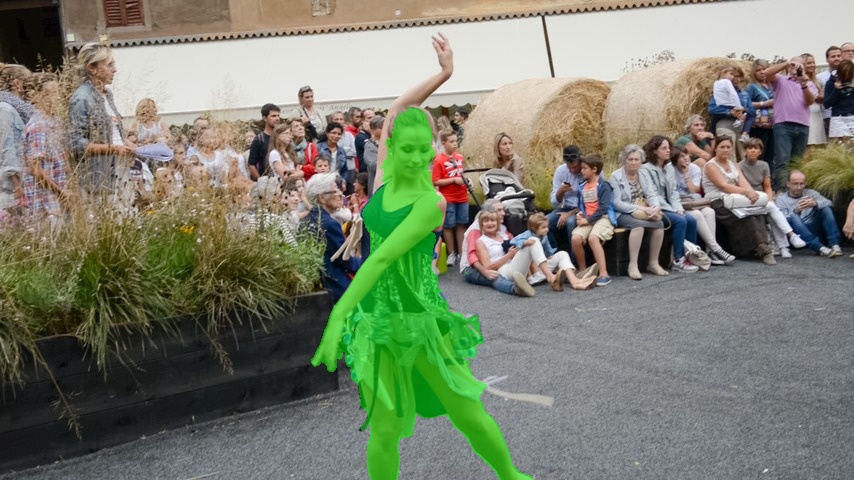}
		\includegraphics[width=0.19\textwidth]{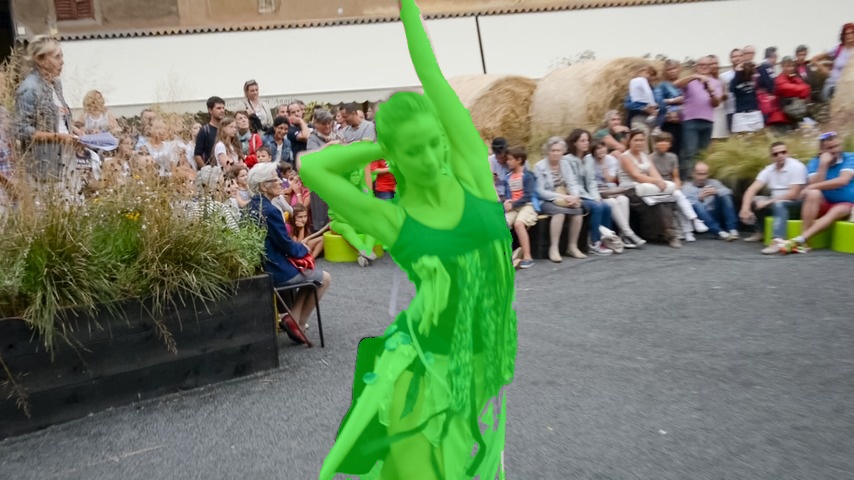}\\
		\includegraphics[width=0.19\textwidth]{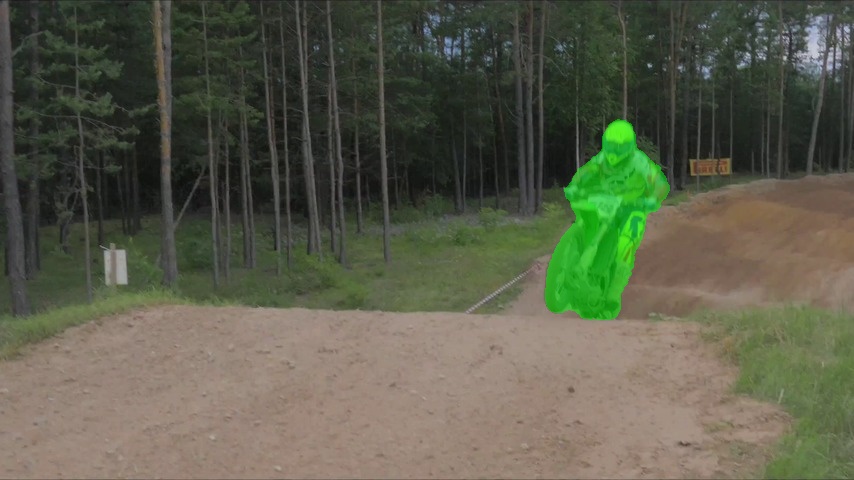}
		\includegraphics[width=0.19\textwidth]{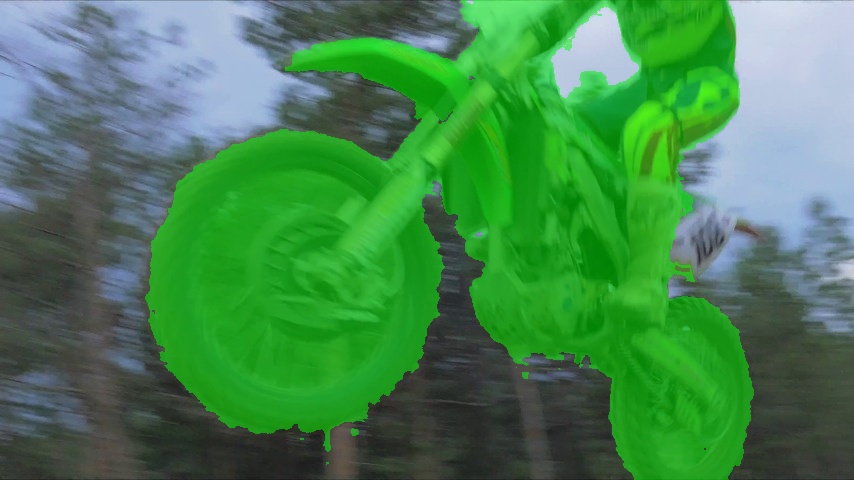}
		\includegraphics[width=0.19\textwidth]{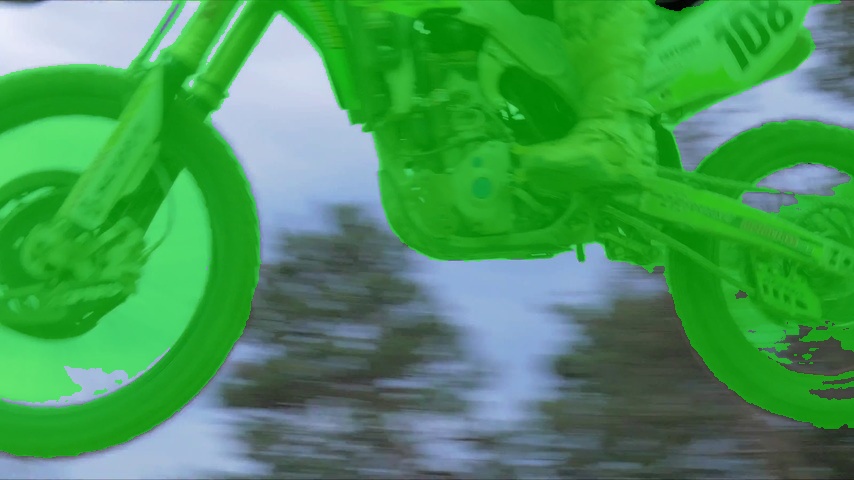}
		\includegraphics[width=0.19\textwidth]{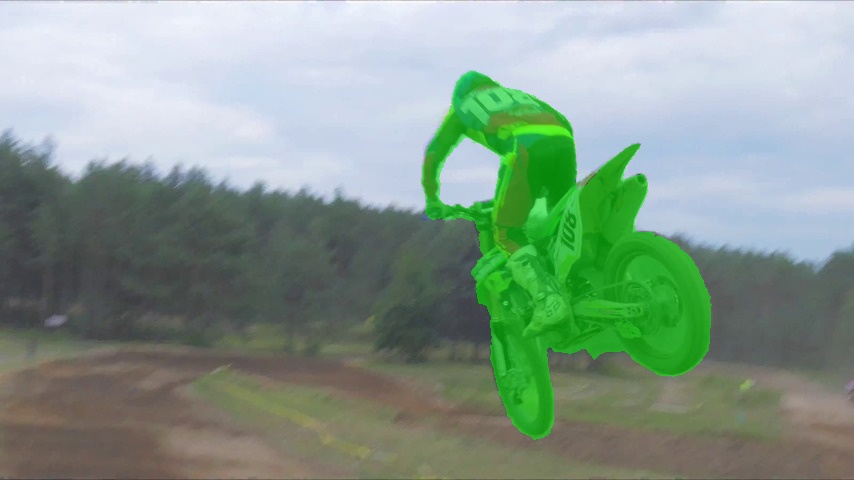}
		\includegraphics[width=0.19\textwidth]{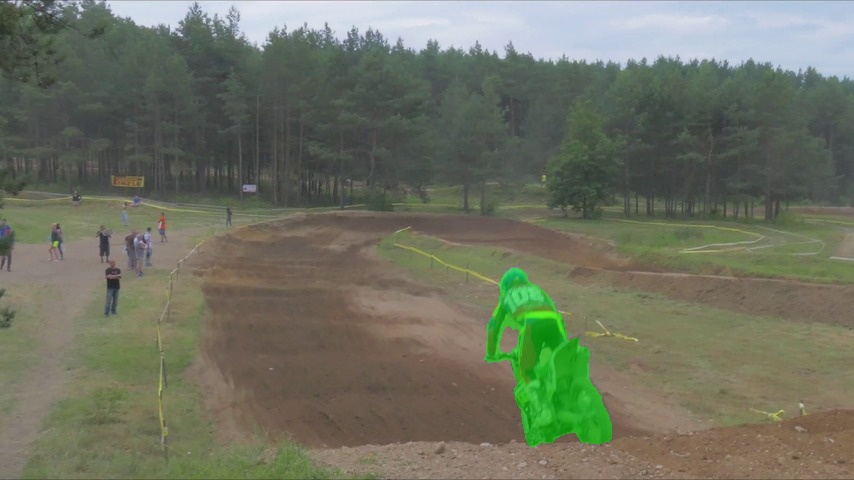}\\
		\includegraphics[width=0.19\textwidth]{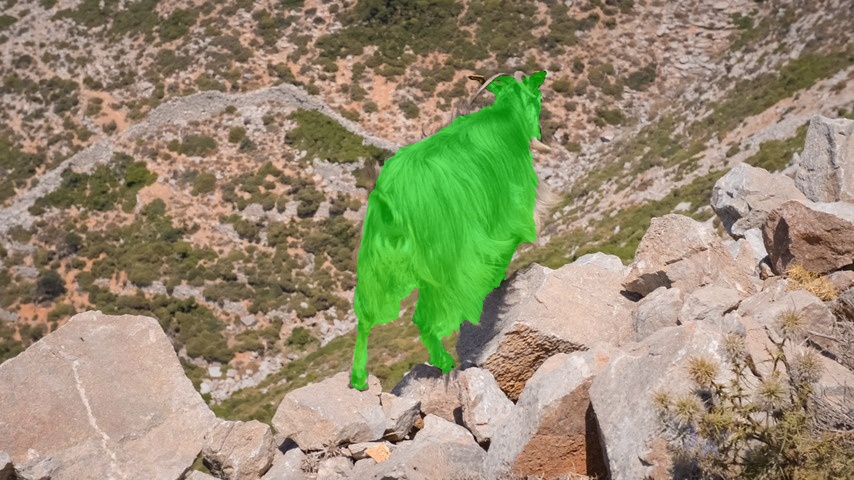}
		\includegraphics[width=0.19\textwidth]{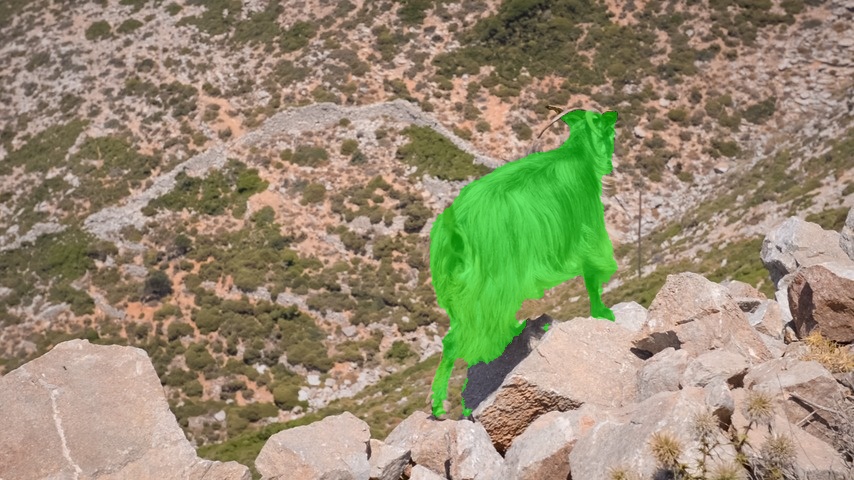}
		\includegraphics[width=0.19\textwidth]{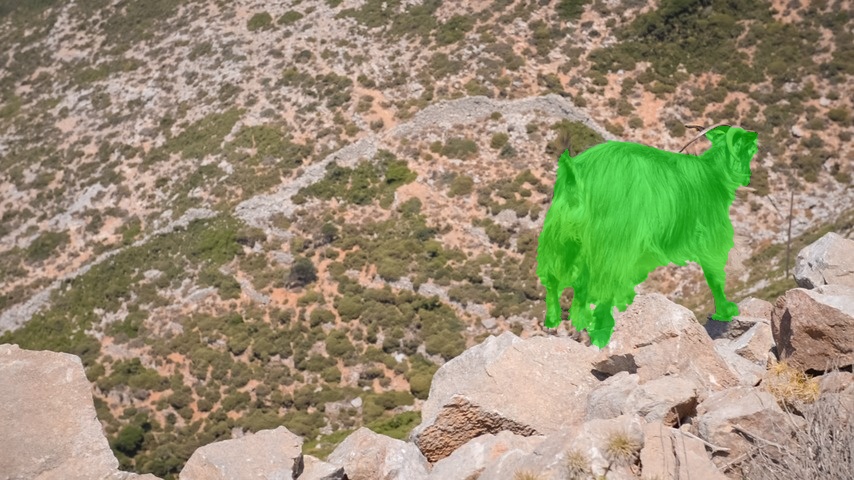}
		\includegraphics[width=0.19\textwidth]{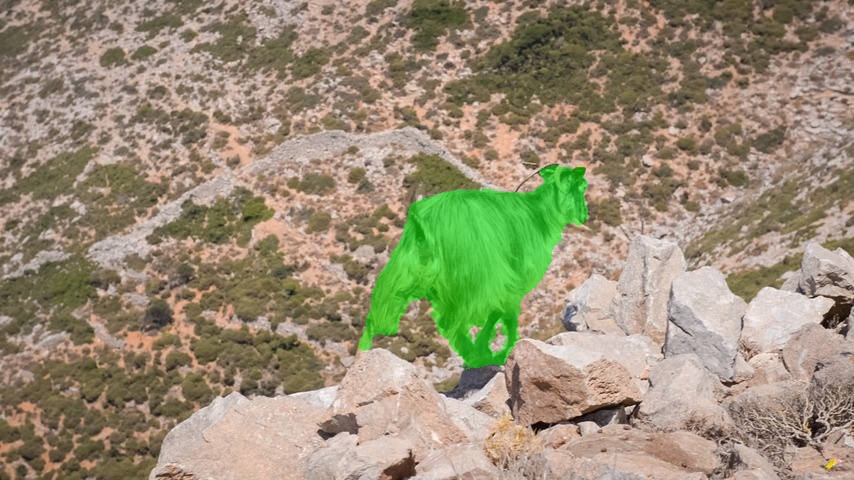}
		\includegraphics[width=0.19\textwidth]{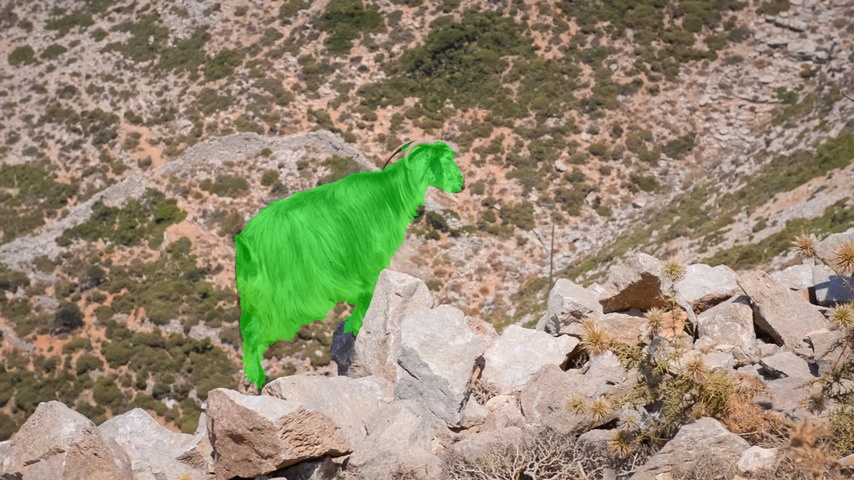}\\
		\vspace{0.1cm}
		\caption{Example qualitative results on DAVIS dataset. Our method performs well on videos with large appearance changes (top row), confusing backgrounds (second row, with people in background), changing viewing angle (third row), and unseen semantic categories (forth row, with a goat as the foreground). Best viewed in color.}\label{fig:example_results_davis}
	\end{figure*}

	\subsection{Datasets and evaluation}
	We evaluate the proposed method on the DAVIS dataset~\cite{Perazzi2016}, Freiburg-Berkeley Motion Segmentation (FBMS) dataset ~\cite{fbms59}, and the SegTrack-v2 dataset~\cite{segtrack}. Note that neither the embedding network nor the optical flow network has been trained on these datasets.
	
	\noindent\textbf{DAVIS.}
	The DAVIS 2016 dataset~\cite{Perazzi2016} is a recently constructed dataset, containing 50 video  sequences in total, with 30 in the \textit{train} set and 20 in the \textit{val} set. It provides binary segmentation ground truth masks for all 3455 frames. This dataset contains challenging videos featuring object deformation, occlusion, and motion blur. The ``target object'' may consist of multiple objects that move together, e.g., a bike with the rider. To evaluate our method, we adopt the protocols in \cite{Perazzi2016}, which include region similarity and boundary accuracy, denoted by $\mathcal{J}$ and $\mathcal{F}$, respectively. $\mathcal{J}$ is computed as the intersection over union (IoU) between the segmentation results and the ground truth. $\mathcal{F}$ is the harmonic mean of boundary precision and recall.
	
	\noindent\textbf{FBMS.} The FBMS dataset~\cite{fbms59} contains 59 video sequences with 720 frames annotated. In contrast to DAVIS, multiple moving objects are annotated separately in FBMS. We convert the instance-level annotations to binary masks by merging all foreground annotations, as in \cite{tokmakov2017learning}. The evaluation metrics include the F-score evaluation protocol proposed in \cite{fbms59} as well as $\mathcal{J}$ used for DAVIS.
	
	\noindent\textbf{SegTrack-v2.}
	The SegTrack-v2 dataset~\cite{segtrack} contains 14 videos with a total of 976 frames. Annotations of individual moving objects are provided for all frames. As with FBMS, the union of the object masks is converted to a binary mask for unsupervised video object segmentation evaluation. For this dataset, we only use $\mathcal{J}$ for evaluation to be consistent with previous work.
	
	\subsection{Implementation details}
	We use the image instance segmentation network trained on the PASCAL VOC 2012 dataset~\cite{fathi2017semantic} to extract the object embedding and objectness. The instance segmentation network is based on DeepLab-v2~\cite{DeepLab} with ResNet~\cite{resnet} as the backbone. We use the stabilized optical flow from a reimplementation of FlowNet2.0~\cite{flownet}. The dimension for the embedding vector, $E$, is 64. The window size $n$ to identify the candidate set $C$ is set to 9 for DAVIS/FBMS, and 5 for SegTrack-v2. For frames in DAVIS dataset, the 9x9 window results in approximately 200 candidates in the embedding edge map. We select $N_S=100$ seeds from the candidates. The initial number of background seeds $N_{BG}$ is $N_S/5$. To add FG seeds as in Sec.~\ref{sec:seed_refinement}, $\alpha$ is set to 0.5. The thresholds for BG seed selection are $O_{BG}=0.3$ and $M_{BG}=0.01$. The CRF parameters are identical with the ones in DeepLab~\cite{DeepLab} (used for the PASCAL dataset). We first tuned all of these parameters on the DAVIS \textit{train} set, where we our $\mathcal{J}$ was 77.5\%. We updated the window size $n$ for SegTrack-v2 empirically, considering the video resolution.
	
	\subsection{Comparing to the state-of-the-art}
	\noindent\textbf{DAVIS.} As shown in Tab.~\ref{table:davis}, we obtain the best performance for unsupervised video object segmentation: 2.3\% higher than the second best and 2.6\% higher than the third best. Our unsupervised approach even outperforms some of the semi-supervised methods that have access to the first frame annotations, VPN~\cite{VPN} and SegFlow~\cite{segflow}, by more than 2\%\footnote{Numeric results for these two methods are listed in Tab.~\ref{table:semi}.}. Some qualitative segmentation results are shown in Fig.~\ref{fig:example_results_davis}.

	\noindent\textbf{FBMS.} We evaluate the proposed method on the \textit{test} set, with 30 sequences in total. The results are shown in Tab.~\ref{table:fbms}. Our method achieves an F-score of 82.8\%: 5.0\% higher than the second best method~\cite{tokmakov2017learning}. Our method's $\mathcal{J}$ mean is more than 10\% better than ARP~\cite{kohprimary}, which performs the second best on DAVIS.

	\begin{table*}[t!h]
		\begin{center}
			\setlength\tabcolsep{5pt}
			\begin{tabular}{|c|c|c|c|c|c|c|c|c|}
				\hline
				&NLC~\cite{NLC} &CUT~\cite{CUT} &FST~\cite{FST} &CVOS\cite{CVOS} &LVO~\cite{tokmakov2017learning} &MP~\cite{LMP} & ARP\cite{kohprimary} & Ours \\
				\hline
				F-score &- &76.8 &69.2 &74.9 &77.8 &77.5 &- &\textbf{82.8}\\
				\hline
				$\mathcal{J}$ Mean &44.5 &-	&55.5 &-  &- &- &59.8 & \textbf{71.9}\\
				\hline
			\end{tabular}
		\end{center}
		\caption{The results on the \textit{test} set of FBMS dataset~\cite{fbms59}. Our method achieves the highest in both evaluation metrics.}
		\label{table:fbms}
		
	\end{table*}

	\noindent\textbf{SegTrack-v2.} We achieve a $\mathcal{J}$ of 59.3\% on this dataset, which is higher than other methods that do well on DAVIS, LVO~\cite{tokmakov2017learning} (57.3\%) and FST~\cite{FST} (54.3\%). Due to low resolution of SegTrack-v2 and the fact that SegTrack-v2 videos can have multiple moving objects of the same class in the background, we are weaker than NLC~\cite{NLC} (67.2\%) in this dataset.
	
	% \begin{table}[t!h]
	% \begin{center}
	% 	\begin{tabular}{|c|c|c|c|c|}
	% 		\hline
	% 		Method  & LVO~\cite{tokmakov2017learning} & FST~\cite{FST} & NLC~\cite{NLC} &Ours\\
	% 		\hline
	% 		$\mathcal{J}$ Mean &57.3 &54.3 &67.2 &59.3\\
	% 		\hline
	% 	\end{tabular}
	% \end{center}
	% \caption{The $\mathcal{J}$ mean on all sequences of the SegTrack-v2~\cite{segtrack} dataset.}
	% \label{table:segtrack}
	% \end{table}

	\subsection{Ablation studies}\label{sec:ablation}
	\noindent\textbf{The effectiveness of instance embedding.} To prove that instance embeddings are more effective than feature embeddings from semantic segmentation networks in our method, we compare against the $fc7$ features from DeepLab-v2~\cite{DeepLab}. % and show the results in Tab.~\ref{table:deeplab_features}. 
	% BS: I think this baseline is confusing here. We could add it to the unsupervised segmentation results table instead.
	%As an additional baseline, we also compare our method against DeeLab-v2 directly trained to produce foreground masks. That network is trained on the PASCAL dataset ~\cite{Pascal2010} with all semantic categories treated as foreground. The PASCAL dataset guides the network to learn general objectness and achieves 65.3\% without any training on DAVIS. 
	Replacing the instance embedding in our method with DeepLab fc7 features achieves 65.2\% in $\mathcal{J}$, more than 10\% less than the instance embedding features. The instance embedding feature vectors are therefore much better suited to linking objects over time and space than semantic segmentation feature vectors. The explicit pixelwise similarity loss (Eq.~\ref{eq:sim_loss}) used to train instance embeddings helps to produce more stable feature vectors than semantic segmentation.
	
	%TODO: move the footnote text correspondingly. (Do it at the end)
	\footnotetext{The numeric results of previous methods are taken from corresponding papers, where oftentimes only one evaluation metric is reported.}
	
	%SIYANG: may say more about leveraging instance embedding here, such as they can be directly transferred.
	
	% \begin{table}[!h]
	% \begin{center}
	% \setlength\tabcolsep{2pt}
	% 	\begin{tabular}{|c|c|c|c|c|}
	% 		\hline
	% 		Method  & DeepLab binary & DeepLab fc7 &Ours &Ours+CRF\\
	% 		\hline
	% 		$\mathcal{J}$ Mean &65.3 &\~65 &75.7 &78.5\\
	% 		\hline
	% 	\end{tabular}
	% \end{center}
	% \caption{Compare the instance embedding to semantic embedding. Experiments conducted on DAVIS \textit{val} set. [NEED UPDATE]}
	% \label{table:deeplab_features}
	% \end{table}
	
	% % BS: If we want to save space, we can just report these numbers in the text.
	% \begin{table}[!h]
	% \begin{center}
	% \setlength\tabcolsep{2pt}
	% 	\begin{tabular}{|c|c|c|c|c|}
	% 		\hline
	% 		Method  & DeepLab fc7 &Ours &Ours+CRF\\
	% 		\hline
	% 		$\mathcal{J}$ Mean &\~65 &75.7 &78.5\\
	% 		\hline
	% 	\end{tabular}
	% \end{center}
	% \caption{Compare the instance embedding to semantic embedding. Experiments conducted on DAVIS \textit{val} set. [NEED UPDATE]}
	% \label{table:deeplab_features}
	% \end{table}

\begin{figure}
		\centering
		\includegraphics[width=0.8\columnwidth]{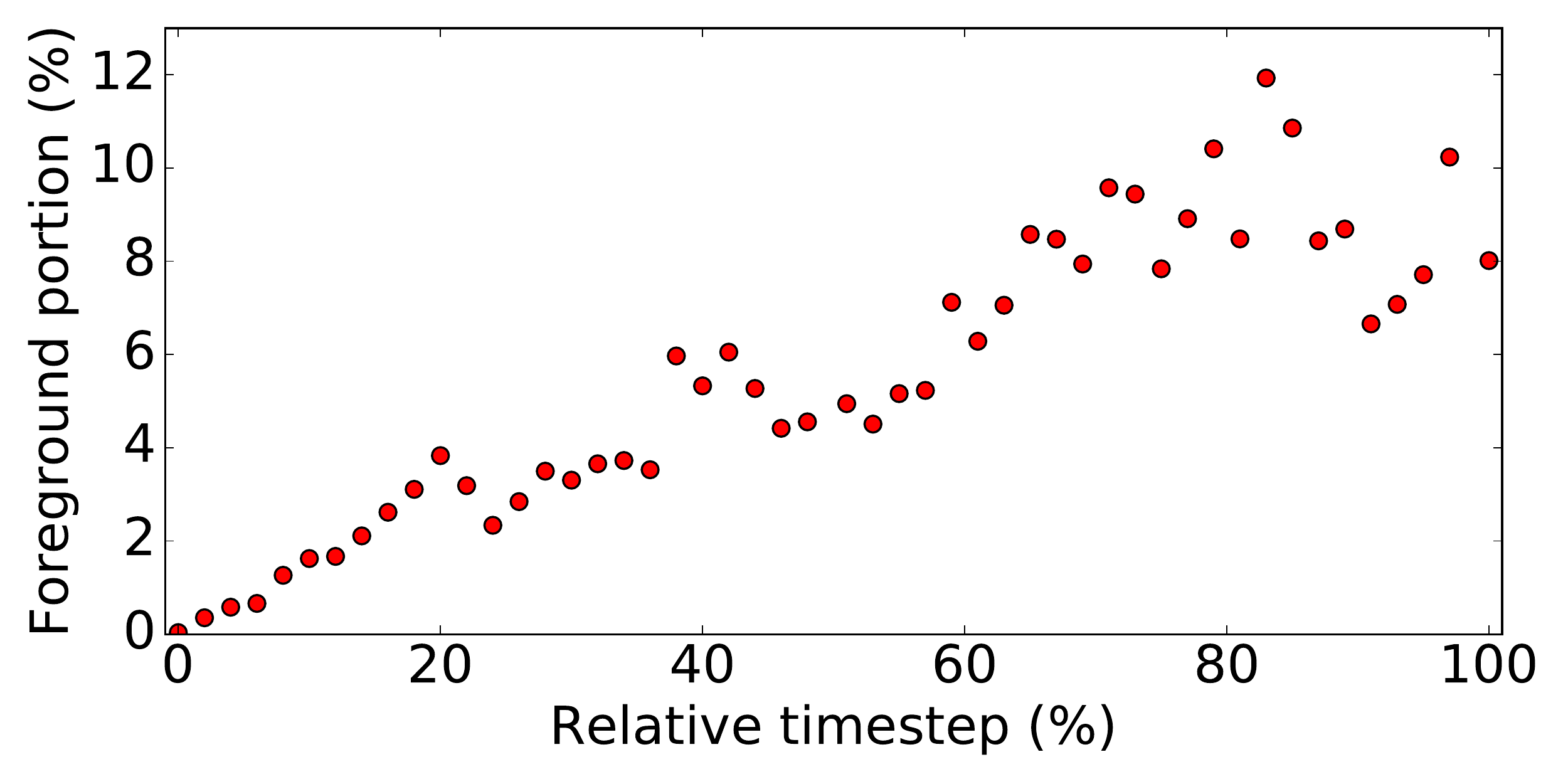}
		\caption{The proportion of incorrectly classified foreground embeddings versus relative timestep. As time progresses, more foreground embeddings are closer to first frame's background than the foreground.}
		\label{fig:temporal}
		\vspace{-0.5cm}
	\end{figure}

	\noindent\textbf{Embedding temporal consistency and online adaptation.}
	We analyze whether embeddings for an object are consistent over time. Given the embeddings for each pixel in every frame and the ground truth foreground masks, we determine how many foreground embeddings in later frames are closer to the background embeddings than foreground embeddings from the first frame. If a foreground embedding from a later frame is closer to any background embedding in the first frame, we call it an incorrectly classified foreground pixel. We plot the proportion of foreground pixels that are incorrectly classified as a function of relative time in the video since the first frame in Fig.~\ref{fig:temporal}. As time from the first frame increases, more foreground embeddings are incorrectly classified. This ``embedding drift'' problem is probably caused by differences in the appearance and location of foreground and background objects in the video.

	To overcome ``embedding drift'', we do online adaptation to update our sets of foreground and background seeds. Updating the seeds is much faster than fine-tuning a neural network to do the adaptation, as done in OnAVOS~\cite{OnAVOS} with a heuristically selected set of examples. The effects of doing online adaptation every $k$ frames are detailed in Tab.~\ref{table:online_adapt}. More frequent online adaptation results in better performance: per-frame online adaptation boosts $\mathcal{J}$ by 7.0\% over non-adaptive seed sets from the first frame.
	
	\begin{table}[!h]
		\begin{center}
			\setlength\tabcolsep{1pt}
			\begin{tabular}{|c|c|c|c|c|c|c|}
				\hline
				Adapt every k frame  & k=1 & k=5 &k=10 &k=20 &k=40 &k=$\infty$\\
				\hline
				$\mathcal{J}$ Mean &77.5 &76.4 &75.9 &75.0 &73.6 &70.5\\
				\hline
			\end{tabular}
		\end{center}
		\caption{The segmentation performance versus online adaptation frequency. Experiments conducted on DAVIS \textit{train} set. Note that k=$\infty$ means no online adaptation.}
		\label{table:online_adapt}
		\vspace{-0.3cm}
	\end{table}
	
	\noindent\textbf{Foreground seed track ranking.}
	In this section, we discuss some variants of foreground seed track ranking. In Eq.~\ref{eq:fg_ranking}, the ranking is based on objectness as well as motion saliency. We analyze three variants: motion saliency alone, objectness alone, and objectness+motion saliency. The results are shown in Tab.~\ref{table:fg_ranking_analysis}. The experiments are conducted on the DAVIS \textit{train} set. The initial FG seed accuracy (second row in Tab.~\ref{table:fg_ranking_analysis}) is evaluated as the proportion of the initial foreground seeds located within the ground truth foreground region. We see that combining motion saliency and objectness results in the best performance, outperforming ``motion alone'' and ``objectness alone'' by 4.0\% and 6.4\%, respectively. Final segmentation performance is consistent with the initial foreground seed accuracy, with the combined ranking outperforming the ``motion alone'' and ``objectness alone'' by 3.2\% and 1.8\%, respectively. The advantage of combining motion and objectness is reported in several previous methods as well~\cite{segflow, FSEG, tokmakov2017learning}. It is interesting to see that using objectness only gives lower initial foreground seed accuracy but higher $\mathcal{J}$ mean than motion only. It is probably because of the different errors the two scores make. When errors selecting foreground seeds occur in ``motion only'' mode, it is more likely that seeds representing ``stuff'' (sky, water, road, etc) are selected as the foreground, but when errors occur in ``objectness only'' mode, incorrect foreground seeds are usually located on static objects in the sequence. In the embedding space, static objects are usually closer in the embedding space to the target object than ``stuff'', so these errors are more forgiving. 
	
	\begin{table}[!h]
		\begin{center}
			\setlength\tabcolsep{3pt}
			\begin{tabular}{|c|c|c|c|}
				\hline
				& Motion & Obj. &Motion + Obj.\\
				\hline
				Init. FG seed acc. &90.6 &88.2 &94.6 \\
				\hline
				$\mathcal{J}$ Mean &74.3 &75.7 &77.5 \\
				\hline
			\end{tabular}
		\end{center}
		\caption{The segmentation performance versus foreground ranking strategy. Experiments are conducted on DAVIS \textit{train} set.}
		\vspace{-0.3cm}
		\label{table:fg_ranking_analysis}
	\end{table}

	\subsection{Semi-supervised video object segmentation}
	We extend the method to semi-supervised video object segmentation by selecting the foreground seeds and background seeds based on the first frame annotation. The seeds covered by the ground truth mask are added to the foreground set $S_{FG}^{0}$ and the rest are added to $S_{BG}^{0}$. Then we apply Eqs.~\ref{eq:distance_compute}-\ref{eq:fg_prob} to all embeddings of the sequence. Results are further refined by a dense CRF. Experiment settings are detailed in supplementary materials. As shown in Tab.~\ref{table:semi}, we achieve 77.6\% in $\mathcal{J}$, better than \cite{segflow} and \cite{VPN}. Note that there are more options for performance improvement such as motion/objectness analysis and online adaptation as we experimented in the unsupervised scenario. We leave those options for future exploration.
	
	\begin{table}[!h]
		\begin{center}
			\setlength\tabcolsep{2pt}
			\begin{tabular}{|c|c|c|c|}
				\hline
				&\small{DAVIS fine-tune?} & \small{Online fine-tune?} &$\mathcal{J}$ Mean\\
				\hline
				OnAVOS~\cite{OnAVOS}&Yes &Yes &86.1\\
				OSVOS~\cite{OSVOS}&Yes &Yes &79.8\\
				SFL~\cite{segflow}&Yes &Yes &76.1\\
				MSK~\cite{MSK} &No &Yes &79.7\\
				VPN~\cite{VPN} &No &No &70.2\\
				\hline
				Ours &No &No &77.6\\
				\hline
			\end{tabular}
		\end{center}
		\caption{The results of semi-supervised video object segmentation on DAVIS \textit{val} set by adopting the proposed method.}
		\label{table:semi}
		\vspace{-0.3cm}
	\end{table}
	
	\section{Conclusion}
	In this paper, we propose a method to transfer the instance embedding learned from static images to unsupervised object segmentation in videos. To be adaptive to the changing foreground in the video object segmentation problem, we train a network to produce embeddings encapsulating instance information rather than training a network that directly outputs a foreground/background score. In the instance embeddings, we identify representative foreground/background embeddings from objectness and motion saliency. Then, pixels are classified based on embedding similarity to the foreground/background. Unlike many previous methods that need to fine-tune on the target dataset, our method achieves the state-of-the-art performance under the unsupervised video object segmentation setting without any fine-tuning, which saves a tremendous amount of labeling effort.

{\small
\bibliographystyle{ieee}
\bibliography{egbib}
}

\newpage
\section{Supplemental materials}
\subsection{Experiment Settings of Semi-supervised Video Object Segmentation}
We extract the $N_S=100$ seeds for the first frame (frame 0) and form image regions, as described in Sec. 3.2 and Sec. 3.3, respectively. Then we compare the image regions with the ground truth mask. For one image region $A_{j}$, if the ground truth mask $G$ covers more than $\alpha$ of $A_{j}$, i.e.,
\begin{align}
|A_{j}\bigcap G| \geq \alpha|A_{j}|,
\end{align}
where $|\cdot|$ denotes the area,
the average embedding within the intersection is computed and added to the foreground embedding set $S^{0}_{FG}$. We set $\alpha = 0.7$. 

\begin{figure}[h]
	\centering
	\includegraphics[width=0.32\columnwidth]{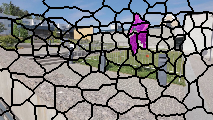}
	\includegraphics[width=0.32\columnwidth]{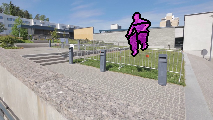}
	\includegraphics[width=0.32\columnwidth]{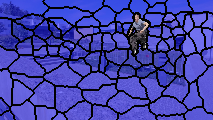}
	\caption{\textbf{Left}: The image regions and the ground truth mask (in magenta) on frame 0.
		\textbf{Center}: The image regions (in magenta) whose average embeddings are used as foreground embeddings for the rest of frames.
		\textbf{Right}: The image regions (in blue) whose average embeddings are used as background embeddings for the rest of frames. Best viewed in color.}
	\label{fig:semi}
\end{figure}

For the background, if $A_{j}$ does not intersect with $G$ at all, i.e.,
\begin{align}
A_{j}\bigcap G = \O,
\end{align}
the average embedding in $A_{j}$ is added to the background embedding set $S^{0}_{BG}$. A visual illustration is shown in Fig.~\ref{fig:semi}.

Then the foreground probability for a pixel on an arbitrary frame is obtained by Eqs. 13-15 and results are further refined by a dense CRF with identical parameters from the unsupervised scenario. We compare our results with multiple previous semi-supervised methods in Tab. 5 in the paper. 
%Most of the mentioned methods have a variant which removes the online fine-tuning, and we list the detailed results in Tab.~\ref{table:semi_more_variants}. Our method achieves the best results in the senario with the time-consuming online fine-tuning prohibited.

% \begin{table}[!h]
% \begin{center}
% \setlength\tabcolsep{2pt}
% \begin{tabular}{|c|c|c|c|c|}
% 	\hline
% 	 &\small{DAVIS FT?} & \small{w/ online FT} & \small{w/o online FT} &$\mathcal{J}$ Mean\\
% 	\hline
% 	OnAVOS~\cite{OnAVOS}&Yes &Yes &86.1\\
% 	OSVOS~\cite{OSVOS}&Yes &Yes &79.8\\
% 	SFL~\cite{segflow}&Yes &Yes &76.1\\
% 	MSK~\cite{MSK} &No &Yes &79.7\\
% 	VPN~\cite{VPN} &No &No &70.2\\
% 	\hline
% 	Ours &No &No &77.6\\
% 	\hline
% \end{tabular}
% \end{center}
% \caption{The results of semi-supervised video object segmentation on DAVIS \textit{val} set by adopting the proposed method.}
% \label{table:semi_more_variants}
% \vspace{-0.3cm}
% \end{table}

\subsection{Per-video Results for DAVIS}
The per-video result are shown for DAVIS \textit{train} set and \textit{val} set are listed in Tab.~\ref{table:train} and Tab.~\ref{table:val}. The evaluation metric is the region similarity $\mathcal{J}$ mentioned in the paper. Note that we used the \textit{train} set for ablation studies (Sec. 4.4), where the masks were \emph{not} refined by dense CRF.

\begin{table}
	\centering
	\setlength\tabcolsep{1pt}
	\begin{tabular}{|c|c|c|c|c|}
		\hline
		Sequence         & ARP~\cite{kohprimary} & FSEG~\cite{FSEG} & Ours & Ours + CRF\\
		\hline
		bear             & 0.92           & 0.907          & 0.935          & \textbf{0.952} \\
		bmx-bumps        & 0.459          & 0.328          & 0.431          & \textbf{0.494} \\
		boat             & 0.436          & \textbf{0.663} & 0.652          & 0.491          \\
		breakdance-flare & 0.815          & 0.763          & 0.809          & \textbf{0.843} \\
		bus              & \textbf{0.849} & 0.825          & 0.848          & 0.842          \\
		car-turn         & 0.87           & 0.903          & \textbf{0.923} & 0.921          \\
		dance-jump       & \textbf{0.718} & 0.612          & 0.674          & 0.716          \\
		dog-agility      & 0.32           & \textbf{0.757} & 0.7            & 0.708          \\
		drift-turn       & 0.796          & \textbf{0.864} & 0.815          & 0.798          \\
		elephant         & 0.842          & \textbf{0.843} & 0.828          & 0.816          \\
		flamingo         & \textbf{0.812} & 0.757          & 0.633          & 0.679          \\
		hike             & \textbf{0.907} & 0.769          & 0.871          & \textbf{0.907} \\
		hockey           & 0.764          & 0.703          & 0.817          & \textbf{0.878} \\
		horsejump-low    & 0.769          & 0.711          & 0.821          & \textbf{0.832} \\
		kite-walk        & 0.599          & 0.489          & 0.598          & \textbf{0.641} \\
		lucia            & 0.868          & 0.773          & 0.863          & \textbf{0.91}  \\
		mallard-fly      & 0.561          & 0.695          & 0.683          & \textbf{0.699} \\
		mallard-water    & 0.583          & 0.794          & 0.76           & \textbf{0.811} \\
		motocross-bumps  & 0.852          & 0.775          & 0.849          & \textbf{0.884} \\
		motorbike        & \textbf{0.736} & 0.407          & 0.685          & 0.708          \\
		paragliding      & \textbf{0.881} & 0.474          & 0.82           & 0.873          \\
		rhino            & \textbf{0.884} & 0.875          & 0.86           & 0.835          \\
		rollerblade      & 0.839          & 0.687          & 0.851          & \textbf{0.896} \\
		scooter-gray     & 0.705          & \textbf{0.733} & 0.686          & 0.655          \\
		soccerball       & 0.824          & 0.797          & 0.849          & \textbf{0.905} \\
		stroller         & \textbf{0.857} & 0.667          & 0.722          & 0.758          \\
		surf             & \textbf{0.939} & 0.881          & 0.87           & 0.902          \\
		swing            & 0.796          & 0.741          & 0.822          & \textbf{0.868} \\
		tennis           & 0.784          & 0.707          & 0.817          & \textbf{0.866} \\
		train            & \textbf{0.915} & 0.761          & 0.766          & 0.765   \\
		\hline
		$\mathcal{J}$ mean & 0.763 & 0.722 & 0.775 & \textbf{0.795} \\
		\hline
	\end{tabular}
	\vspace{0.1cm}
	\caption{Per-video results on DAVIS \textit{train} set. The region similarity $\mathcal{J}$ is reported.}
	\label{table:train}
\end{table}

\begin{table}
	\centering
	\setlength\tabcolsep{1pt}
	\begin{tabular}{|c|c|c|c|c|}
		\hline
		Sequence         & ARP~\cite{kohprimary} & FSEG~\cite{FSEG} & Ours & Ours + CRF\\
		\hline
		blackswan          & \textbf{0.881} & 0.812          & 0.715          & 0.786          \\
		bmx-trees          & \textbf{0.499} & 0.433          & 0.496          & \textbf{0.499} \\
		breakdance         & \textbf{0.762} & 0.512          & 0.485          & 0.555          \\
		camel              & 0.903          & 0.836          & 0.902          & \textbf{0.929} \\
		car-roundabout     & 0.816          & \textbf{0.907} & 0.88           & 0.88           \\
		car-shadow         & 0.736          & 0.896          & 0.929          & \textbf{0.93}  \\
		cows               & 0.908          & 0.869          & 0.91           & \textbf{0.925} \\
		dance-twirl        & \textbf{0.798} & 0.704          & 0.781          & 0.797          \\
		dog                & 0.718          & 0.889          & 0.906          & \textbf{0.936} \\
		drift-chicane      & \textbf{0.797} & 0.596          & 0.76           & 0.775          \\
		drift-straight     & 0.715          & 0.811          & \textbf{0.884} & 0.857          \\
		goat               & 0.776          & 0.83           & 0.861          & \textbf{0.864} \\
		horsejump-high     & 0.838          & 0.652          & 0.794          & \textbf{0.851} \\
		kite-surf          & 0.591          & 0.392          & 0.569          & \textbf{0.647} \\
		libby              & 0.654          & 0.584          & 0.686          & \textbf{0.743} \\
		motocross-jump     & \textbf{0.823} & 0.775          & 0.754          & 0.778          \\
		paragliding-launch & 0.601          & 0.571          & 0.595          & \textbf{0.624} \\
		parkour            & 0.828          & 0.76           & 0.852          & \textbf{0.909} \\
		scooter-black      & \textbf{0.746} & 0.688          & 0.727          & 0.74           \\
		soapbox            & \textbf{0.846} & 0.624          & 0.668          & 0.67\\
		\hline
		$\mathcal{J}$ mean & 0.762	&0.707	&0.758	&\textbf{0.785} \\
		\hline
	\end{tabular}
	\vspace{0.1cm}
	\caption{Per-video results on DAVIS \textit{val} set. The region similarity $\mathcal{J}$ is reported.}
	\label{table:val}
\end{table}

\subsection{Instance Embedding Drifting}
\begin{figure}
	\centering
	\includegraphics[width=\columnwidth]{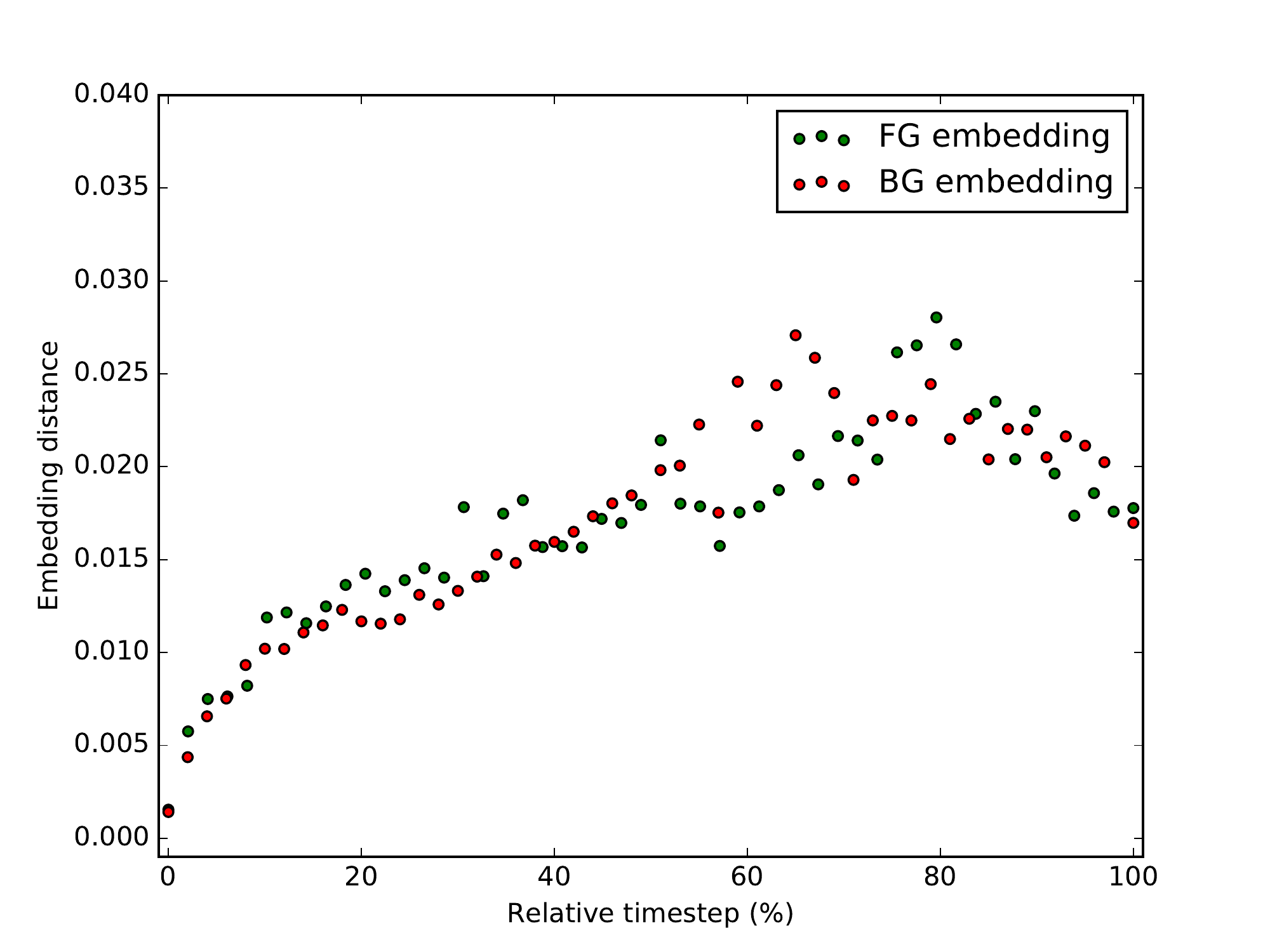}
	\caption{The FG/BG distance between later frames and frame 0. Both FG/BG embeddings become farther from their reference embedding on frame 0.}
	\label{fig:drift}
\end{figure}

In Sec.~4.4 of the paper, we mentioned the ``embedding drift'' problem. Here we conduct another experiment to demonstrate that the embedding changes gradually with time.
In this experiment, we extract foreground and background embeddings based on the ground truth masks for every frame. The embeddings from the first frame (frame 0) are used as references. We compute the average distance between the foreground/background embeddings from an arbitrary frame and the reference embeddings. Mathematically,
\begin{align}
d_{FG}(k, 0) = \frac{1}{|FG_{k}|}\sum_{j\in FG^{k}}\min_{l \in FG^{0}}||\mathbf{f}(j)-\mathbf{f}(l)||_{2},\\
d_{BG}(k, 0) = \frac{1}{|BG_{k}|}\sum_{j\in BG^{k}}\min_{l \in BG^{0}}||\mathbf{f}(j)-\mathbf{f}(l)||_{2},
\end{align}
where $FG^{k}$ and $BG_{k}$ denote the ground truth foreground and background regions, respectively, $\mathbf{f}(j)$ denotes the embedding corresponding to pixel $j$, and $d_{FG}(k, 0)$/$d_{BG}(k, 0)$ represent the foreground/background embedding distance between frame $k$ and frame 0. Then we average $d_{FG}(k, 0)$ and $d_{BG}(k, 0)$ across sequences and plot their relationship with the relative timestep in Fig.~\ref{fig:drift}. As we observe, the embedding distance is increasing with time elapsing. Namely, both objects and background become less similar to themselves on frame 0, which supports the necessity of online adaptation.
% Boundary embedding analysis

\subsection{More visual examples}
\begin{figure*}
	\centering
	\includegraphics[width=0.19\textwidth]{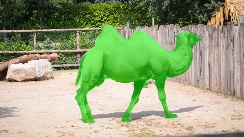}
	\includegraphics[width=0.19\textwidth]{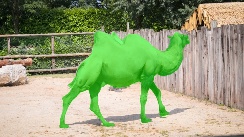}
	\includegraphics[width=0.19\textwidth]{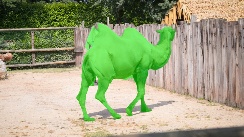}
	\includegraphics[width=0.19\textwidth]{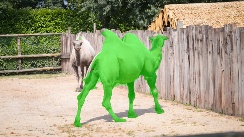}
	\includegraphics[width=0.19\textwidth]{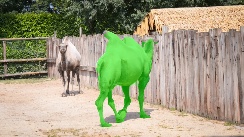}\\
	\includegraphics[width=0.19\textwidth]{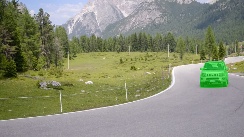}
	\includegraphics[width=0.19\textwidth]{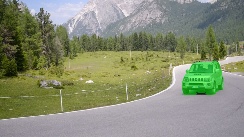}
	\includegraphics[width=0.19\textwidth]{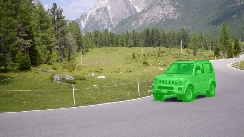}
	\includegraphics[width=0.19\textwidth]{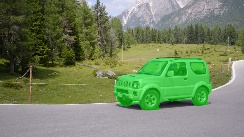}
	\includegraphics[width=0.19\textwidth]{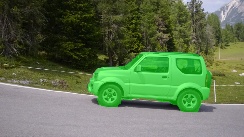}\\
	\includegraphics[width=0.19\textwidth]{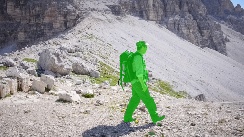}
	\includegraphics[width=0.19\textwidth]{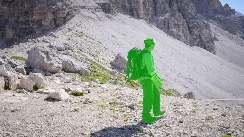}
	\includegraphics[width=0.19\textwidth]{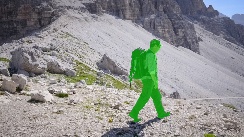}
	\includegraphics[width=0.19\textwidth]{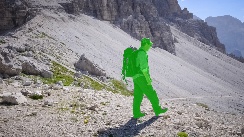}
	\includegraphics[width=0.19\textwidth]{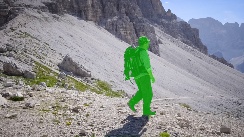}\\
	\includegraphics[width=0.19\textwidth]{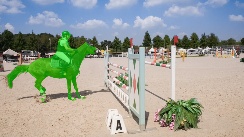}
	\includegraphics[width=0.19\textwidth]{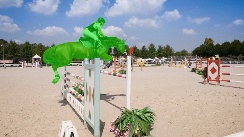}
	\includegraphics[width=0.19\textwidth]{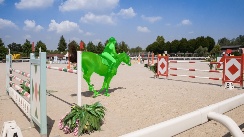}
	\includegraphics[width=0.19\textwidth]{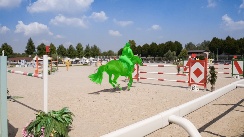}
	\includegraphics[width=0.19\textwidth]{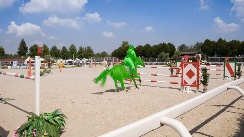}\\
	\includegraphics[width=0.19\textwidth]{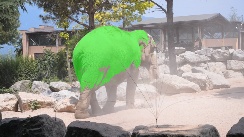}
	\includegraphics[width=0.19\textwidth]{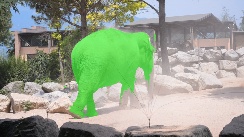}
	\includegraphics[width=0.19\textwidth]{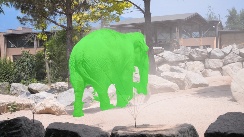}
	\includegraphics[width=0.19\textwidth]{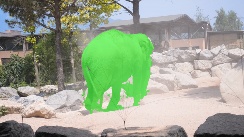}
	\includegraphics[width=0.19\textwidth]{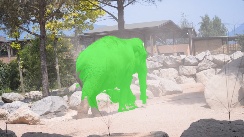}\\
	\includegraphics[width=0.19\textwidth]{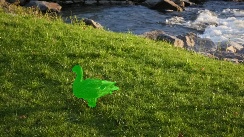}
	\includegraphics[width=0.19\textwidth]{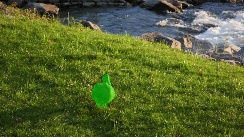}
	\includegraphics[width=0.19\textwidth]{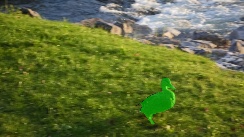}
	\includegraphics[width=0.19\textwidth]{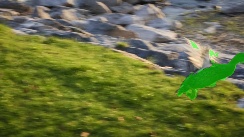}
	\includegraphics[width=0.19\textwidth]{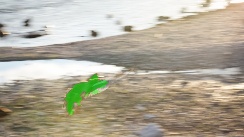}\\
	\includegraphics[width=0.19\textwidth]{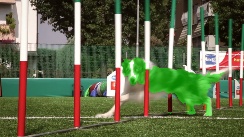}
	\includegraphics[width=0.19\textwidth]{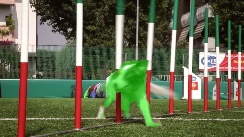}
	\includegraphics[width=0.19\textwidth]{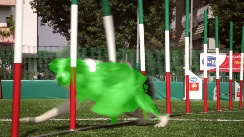}
	\includegraphics[width=0.19\textwidth]{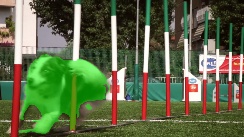}
	\includegraphics[width=0.19\textwidth]{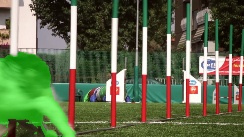}\\
	\includegraphics[width=0.19\textwidth]{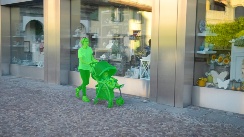}
	\includegraphics[width=0.19\textwidth]{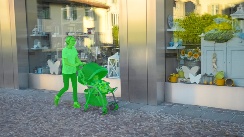}
	\includegraphics[width=0.19\textwidth]{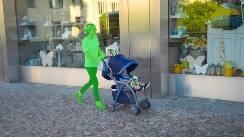}
	\includegraphics[width=0.19\textwidth]{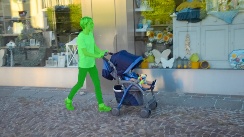}
	\includegraphics[width=0.19\textwidth]{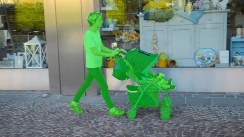}\\
	\includegraphics[width=0.19\textwidth]{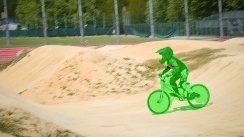}
	\includegraphics[width=0.19\textwidth]{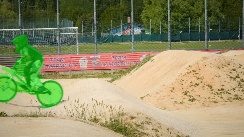}
	\includegraphics[width=0.19\textwidth]{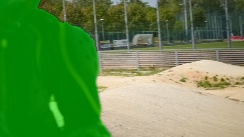}
	\includegraphics[width=0.19\textwidth]{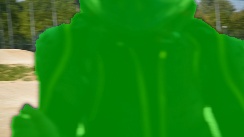}
	\includegraphics[width=0.19\textwidth]{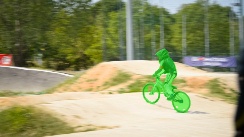}\\
	\includegraphics[width=0.19\textwidth]{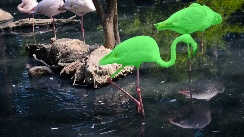}
	\includegraphics[width=0.19\textwidth]{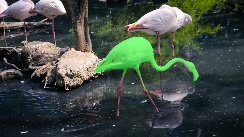}
	\includegraphics[width=0.19\textwidth]{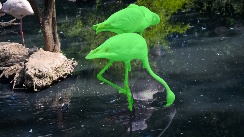}
	\includegraphics[width=0.19\textwidth]{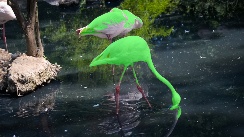}
	\includegraphics[width=0.19\textwidth]{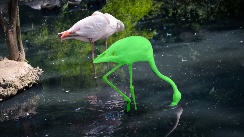}
	\caption{Visual examples from the DAVIS dataset. The ``camel'' sequence (first row) is mentioned as an example where the static camel (the one not covered by our predicted mask) acts as hard negatives because it is semantically similar with foreground while belongs to the background. Our method correctly identifies it as background from motion saliency. The last three rows show some failure cases. In the "stroller" sequence (third last row), our method fails to include the stroller for some frames. In the "bmx-bump" sequence (second last row), when the foreground, namely the rider and the bike, is totally occluded, our method wrongly identifies the occluder as foreground. The ``flamingo'' sequence (last row) illustrates a similar situation with the ``camel'' sequence, where the proposed method does less well due to imperfect optical flow (the foreground mask should include only the flamingo located in the center of each frame). Best viewed in color.}
	\label{fig:DAVIS}
\end{figure*}

\begin{figure*}[!p]
	\includegraphics[width=0.19\textwidth]{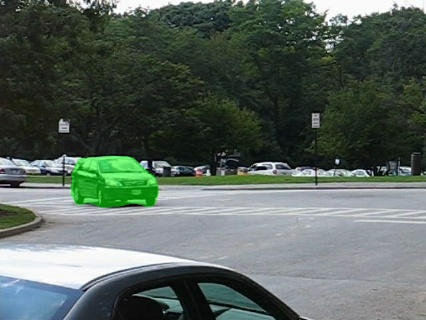}
	\includegraphics[width=0.19\textwidth]{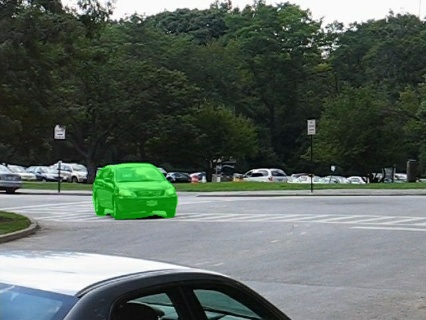}
	\includegraphics[width=0.19\textwidth]{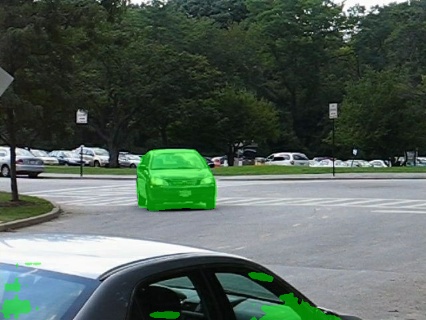}
	\includegraphics[width=0.19\textwidth]{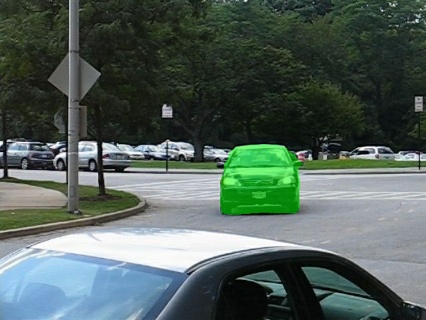}
	\includegraphics[width=0.19\textwidth]{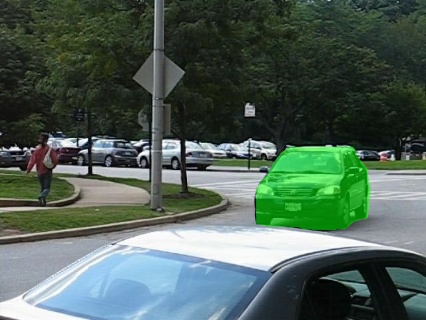}\\
	\includegraphics[width=0.19\textwidth]{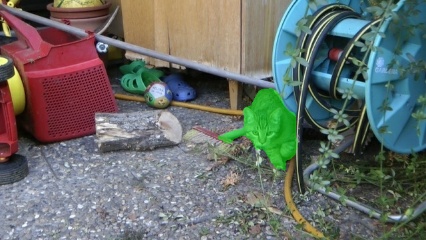}
	\includegraphics[width=0.19\textwidth]{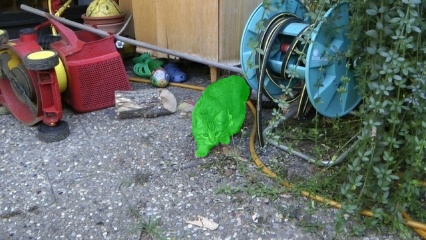}
	\includegraphics[width=0.19\textwidth]{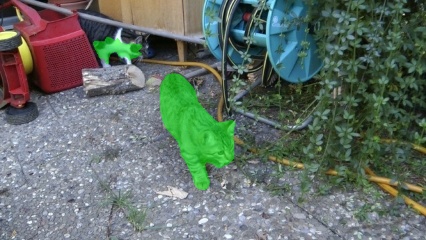}
	\includegraphics[width=0.19\textwidth]{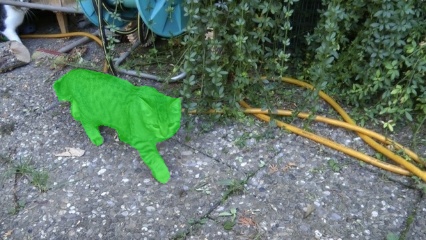}
	\includegraphics[width=0.19\textwidth]{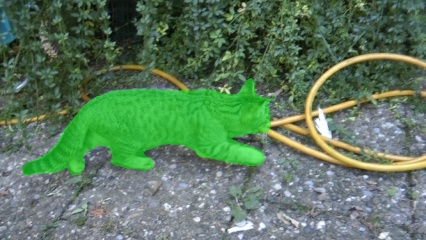}\\
	\includegraphics[width=0.19\textwidth]{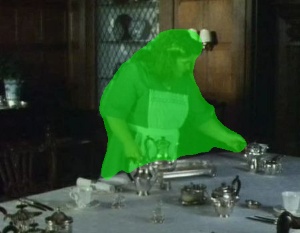}
	\includegraphics[width=0.19\textwidth]{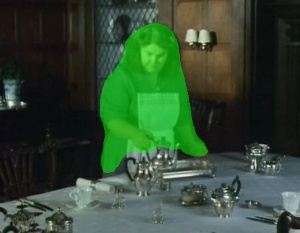}
	\includegraphics[width=0.19\textwidth]{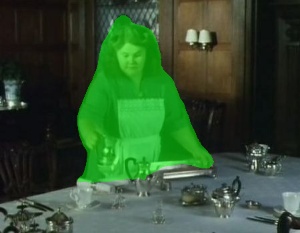}
	\includegraphics[width=0.19\textwidth]{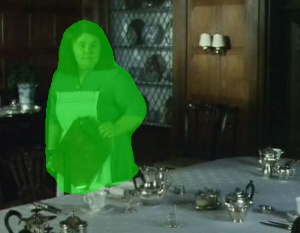}
	\includegraphics[width=0.19\textwidth]{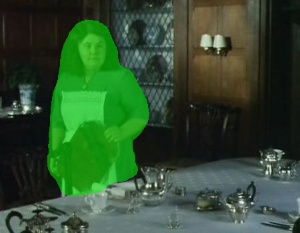}\\
	\includegraphics[width=0.19\textwidth]{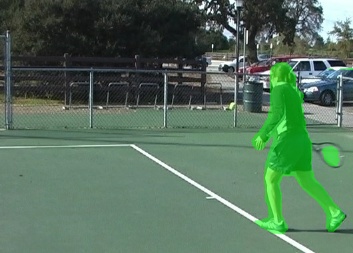}
	\includegraphics[width=0.19\textwidth]{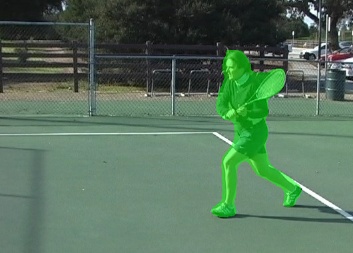}
	\includegraphics[width=0.19\textwidth]{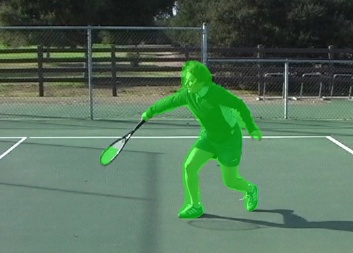}
	\includegraphics[width=0.19\textwidth]{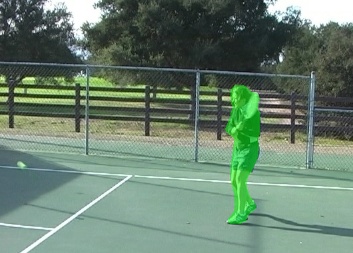}
	\includegraphics[width=0.19\textwidth]{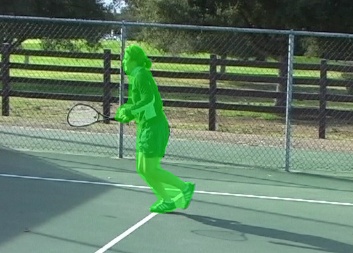}\\
	\includegraphics[width=0.19\textwidth]{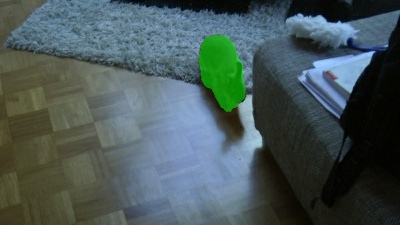}
	\includegraphics[width=0.19\textwidth]{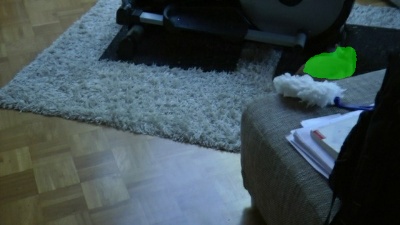}
	\includegraphics[width=0.19\textwidth]{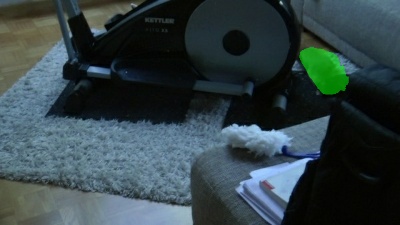}
	\includegraphics[width=0.19\textwidth]{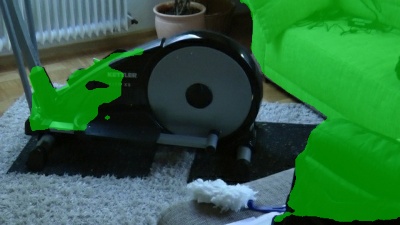}
	\includegraphics[width=0.19\textwidth]{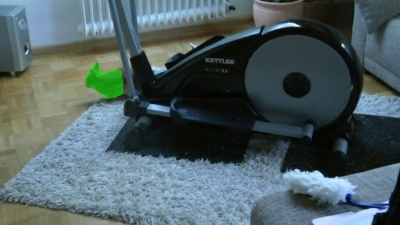}\\
	\includegraphics[width=0.19\textwidth]{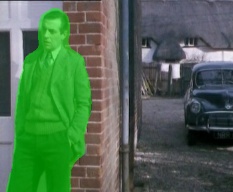}
	\includegraphics[width=0.19\textwidth]{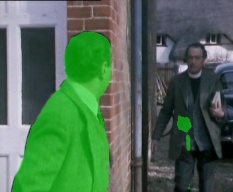}
	\includegraphics[width=0.19\textwidth]{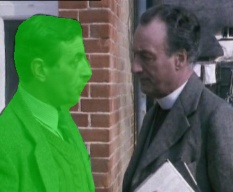}
	\includegraphics[width=0.19\textwidth]{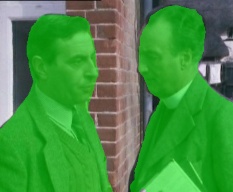}
	\includegraphics[width=0.19\textwidth]{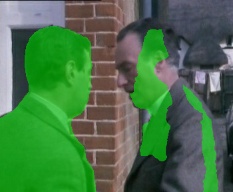}
	\caption{Visual examples from the FBMS dataset. The last two rows show some failure cases. In the ``rabbits04'' sequence (second last row), the foreground is wrongly identified when the rabbit is wholy occluded. In the ``marple6'' sequence (last row), the foreground should include two people, but our method fails on some frames because one of them demonstrates low motion saliency. Best viewed in color.}
	\label{fig:FBMS}
\end{figure*}

We provide more visual examples for the DAVIS dataset~\cite{Perazzi2016} and the FBMS dataset~\cite{fbms59} in Fig.~\ref{fig:DAVIS} and Fig.~\ref{fig:FBMS}, respectively. Furthermore, the results for all annotated frames in DAVIS and FBMS are attached in the folders named ``DAVIS'' and ``FBMS'' submitted together with this document. The frames are resized due to size limit.

\end{document}